\pdfoutput=1

\documentclass[11pt]{article}

\usepackage{fontawesome5}

\usepackage[final]{acl}

\usepackage{times}
\usepackage{latexsym}
\usepackage{booktabs}
\usepackage{adjustbox}

\usepackage[T1]{fontenc}

\usepackage[utf8]{inputenc}

\usepackage{microtype}

\usepackage{inconsolata}

\usepackage{graphicx}
\usepackage{fontawesome5} 

\usepackage{tcolorbox}
\usepackage{multirow}  
\usepackage{rotating}
\usepackage{subcaption}
\usepackage{float}

\title{skLEP: A Slovak General Language Understanding Benchmark}

\author{\bf Marek \v{S}uppa$^{\alpha,~\beta}$ \quad Andrej Ridzik$^{\delta}$ \quad Daniel Hl\'adek$^{\gamma}$ \quad Tom\'{a}\v{s} Jav\r{u}rek$^{\delta}$\\ 
\bf Vikt\'{o}ria Ondrejov\'{a}$^{\alpha,~\beta}$ \quad Krist\'{\i}na S\'{a}sikov\'{a}$^{\alpha}$ \quad Martin Tamajka$^{\delta}$ \quad Mari\'an \v{S}imko$^{\delta}$\\
$^{\alpha}$Comenius University in Bratislava, Slovakia, 
$^{\beta}$Cisco Systems, \\
$^{\gamma}$Technical University of Košice, Slovakia, \\
$^{\delta}$Kempelen Institute of Intelligent Technologies, Bratislava, Slovakia \\
{\small \textbf{Correspondence:} marek@suppa.sk}
}

\begin{document}
\maketitle
\begin{abstract}
In this work, we introduce skLEP, the first comprehensive benchmark specifically designed for evaluating Slovak natural language understanding (NLU) models. We have compiled skLEP to encompass nine diverse tasks that span token-level, sentence-pair, and document-level challenges, thereby offering a thorough assessment of model capabilities. To create this benchmark, we curated new, original datasets tailored for Slovak and meticulously translated established English NLU resources. Within this paper, we also present the first systematic and extensive evaluation of a wide array of Slovak-specific, multilingual, and English pre-trained language models using the skLEP tasks. Finally, we also release the complete benchmark data, an open-source toolkit facilitating both fine-tuning and evaluation of models, and a public leaderboard at \url{https://github.com/slovak-nlp/sklep} in the hopes of fostering reproducibility and drive future research in Slovak NLU.
\end{abstract}

\section{Introduction}
The field of Natural Language Processing (NLP) has shifted towards large pre-trained models
capable of handling a wide range of tasks. This trend has underscored the need for a
standardized evaluation suite to test models across diverse tasks. Benchmarks such as GLUE
\cite{wang2018glue} and SuperGLUE \cite{wang2019superglue} have become widely adopted,
driving the development of models that excel on these tests and in general natural
language understanding \cite{devlin2018bert,liu2019roberta}.

However, these benchmarks primarily focus on English. In response, recent work has
developed benchmarks for individual non-English languages
\cite{augustyniak2022way, le2019flaubert, shavrina2020russiansuperglue} and for multilingual
evaluation \cite{xu-etal-2020-clue, liang2020xglue, hu2020xtreme, ruder2021xtreme}. Yet, this
development remains skewed toward high-resource languages, often overlooking low- to
mid-resource ones.

In this work, we introduce \textbf{skLEP}, a GLUE-style benchmark for Slovak, a mid-resource language \cite{joshi-etal-2020-state} with 10 million native speakers, comprising nine
tasks. Although similar benchmarks exist for other Slavic languages such as Bulgarian
\cite{hardalov-etal-2023-bgglue}, Polish \cite{rybak2020klej}, Russian
\cite{shavrina2020russiansuperglue}, and Slovene \cite{vzagar2022slovene}, no equivalent has
been established for Slovak. This is especially problematic given the emergence of several
Slovak-specific large language models \cite{pikuliak2021slovakbert,FERNETC5,drzik2024slovak,
de-gibert-etal-2024-new-massive}, making it difficult to benchmark their performance
against well-established multilingual alternatives.

Creating a GLUE-style benchmark for Slovak posed non-trivial challenges. While high-quality
datasets exist for various tasks, they do not cover the full range expected in such a benchmark.
Thus, we introduced new datasets and translated established English resources to compile the
complete set of tasks in skLEP.

Using the tasks in skLEP, we conduct what is, to the best of our knowledge, the first systematic
evaluation of existing language models on Slovak by finetuning them on our benchmark tasks and
comparing their performance across the entire suite.

\begin{table*}[h!]
    \centering
    \resizebox{\textwidth}{!}{
    \begin{tabular}{clcccccl}
        \toprule
        \# & \textbf{Corpus} & \textbf{|Train|} & \textbf{|Dev|} & \textbf{|Test|} &  \textbf{Task} & \textbf{Metric} & \textbf{Domain} \\
        \midrule
        \multicolumn{8}{l}{\textit{Token-Level Tasks}} \\
        \midrule
        1 & \textbf{UD} & 8,483 & 1,060 & 1,061 & POS Tagging & Macro F1 & misc. \\
        2 & \textbf{UNER} & 8,483 & 1,060 & 1,061  & Named Entity Recognition & Macro F1 & misc. \\
        3 & \textbf{WGSK} & 4,687 & 669 & 1,340 &  Named Entity Recognition & Macro F1 & Wikipedia \\
        \midrule
        \multicolumn{8}{l}{\textit{Sentence-Pair Tasks}} \\
        \midrule
        4 & \underline{\textbf{RTE}} & 2,490 & 277 & \underline{1,660}  & Textual Entailment & Accuracy & news, Wikipedia \\
        5 & \underline{\textbf{NLI}} & 392,702 & 2,490 & 5,004 &  NLI & Accuracy & misc. \\
        6 & \underline{\textbf{STS}} & 5,604 & 1,481 & 1,352 &  Semantic Textual Similarity & Pearson Corr. & misc. \\
        \midrule
        \multicolumn{8}{l}{\textit{Document-Level Tasks}} \\
        \midrule
        7 & \underline{\textbf{HS}} & \underline{10,531} & \underline{1,339} & 1,319 & Hate Speech Classification & Accuracy & social media \\
        8 & \textbf{SA} & 3,560 & 522 & 1,042 & Sentiment Analysis & Accuracy & customer reviews \\
        9 & \textbf{QA} & \underline{71,999} & \underline{9,583} & 9,583 &  Question Answering & Macro F1 & Wikipedia \\
        \bottomrule
    \end{tabular}
    }
    \caption{Summary of the tasks that comprise the \textit{skLEP} benchmark. The numbers in the train, development, and test columns represent the number of ''atomic units'' the task features, that is the number of sentences for token classification tasks, number of sentence pairs for the sentence-pair tasks and number of documents for the Document Classification Tasks. The remaining columns briefly describe the task as well as the metric it uses for evaluation. The domain reflects the dataset's original data source. The \underline{underlined} datasets (RTE, NLI, STS and HS) and splits (in HS and QA) are newly introduced in this work.}
\end{table*}

Our contributions can be summarized as follows:

\begin{itemize}
  \item We introduce skLEP, the first GLUE‐style benchmark dedicated to Slovak natural language understanding, spanning token-level, sentence-pair, and document-level tasks.
  \item We compile nine diverse tasks by curating new datasets and translating established English resources—with native speaker post‐editing—to ensure high-quality evaluation. 
  \item We provide extensive baseline evaluations over Slovak-specific, multilingual, and English models using a rigorous hyperparameter grid search, establishing robust performance benchmarks.
  \item We release\footnote{The skLEP code, data and models are available at \url{https://github.com/slovak-nlp/sklep}} an open-source toolkit integrated with the HuggingFace framework and a standardized leaderboard to foster reproducibility and drive future research in Slovak NLU. 
\end{itemize}

\begin{table*}[ht]
\centering
\begin{tabular}{p{0.2cm} l}
\toprule

\parbox[t]{0.2cm}{\rotatebox{90}{\textbf{UD}}} & \parbox{15cm}{
\textbf{Document:} \textit{Je potrebné chrániť bohatstvo lokálnych stredoamerických odrôd kukurice , pretože predstavujú zdroj biodiverzity pre ďalšie jej šľachtenie .} \\
\underline{\textbf{Tags:}} \underline{AUX}, \underline{ADJ}, \underline{VERB}, \underline{NOUN}, \underline{ADJ}, \underline{ADJ}, \underline{NOUN}, \underline{NOUN}, \underline{PUNCT}, \underline{SCONJ}, \underline{VERB}, \underline{NOUN}, \underline{NOUN}, \underline{ADP}, \underline{ADJ}, \underline{DET}, \underline{NOUN}, \underline{PUNCT}
} \\

\midrule

\parbox[t]{0.2cm}{\rotatebox{90}{\textbf{UNER}}} & \parbox{15cm}{\textbf{Document:} \textit{V pakte medzi Hitlerom a Stalinom bolo Fínsko pridelené do sféry ZSSR .} \\
\underline{\textbf{Tags:}} \underline{O}, \underline{O}, \underline{O}, \underline{B-PER}, \underline{O}, \underline{B-PER}, \underline{O}, \underline{B-LOC}, \underline{O}, \underline{O}, \underline{O}, \underline{B-ORG}, \underline{O}
} \\

\midrule

\parbox[t]{0.2cm}{\rotatebox{90}{\textbf{WGSK}}} & \parbox{15cm}{
\textbf{Document:} \textit{Počas druhej svetovej vojny tu od roku 1939 do roku 1945 boli uskladnené umelecké zbierky Parížskeho múzea v Louvre .} \\
\underline{\textbf{Tags:}} \underline{O}, \underline{B-MISC}, \underline{I-MISC}, \underline{I-MISC}, \underline{O}, \underline{O}, \underline{O}, \underline{O}, \underline{O}, \underline{O}, \underline{O}, \underline{O}, \underline{O}, \underline{O}, \underline{O}, \underline{B-ORG}, \underline{I-ORG}, \underline{I-ORG}, \underline{I-ORG}, \underline{O}} \\

\midrule

\parbox[t]{0.2cm}{\rotatebox{90}{\textbf{RTE}}} & \parbox{15cm}{
\textbf{Text1:} \textit{Obrúsky, pozvánky a obyčajný starý papier stoja viac ako pred mesiacom.} \\
\textbf{Text2:} \textit{Cena papiera rastie.} \\
\underline{\textbf{Correct Label:}} \underline{Entailment}} \\

\midrule

\parbox[t]{0.2cm}{\rotatebox{90}{\textbf{NLI}}} & \parbox{15cm}{
\textbf{Premise:} \textit{Záblesky múdrosti by sa nemali prehliadať.} \\
\textbf{Hypothesis:} \textit{Záblesky múdrosti nie sú dôležité.} \\
\underline{\textbf{Entailment:}} \underline{Contradiction}} \\

\midrule

\parbox[t]{0.2cm}{\rotatebox{90}{\textbf{STS}}} & \parbox{15cm}{
\textbf{Premise:} \textit{Malý pes leží na posteli.} \\
\textbf{Hypothesis:} \textit{Na posteli leží malý pes.} \\
\underline{\textbf{Similarity Score:}} \underline{5.0}} \\

\midrule

\parbox[t]{0.2cm}{\rotatebox{90}{\textbf{HS}}} & \parbox{15cm}{
\textbf{Text:} \textit{Žiadna vláda kde budú feťáci a narkomani nebude nikdy dobrá} \\
\underline{\textbf{Correct Label:}} \underline{Hate Speech}} \\

\midrule

\parbox[t]{0.2cm}{\rotatebox{90}{\textbf{SA}}} & \parbox{15cm}{
\textbf{Text:} \textit{Pri vstupe do predajne Vás víta príjemný personál, čo mňa presvedčí o tom, že sem treba sa vracať aj druhýkrát, kde človek načerpá novú energiu do seba a samozrejme do svojho auta.} \\
\underline{\textbf{Sentiment:}} \underline{Positive}} \\

\midrule

\parbox[t]{0.2cm}{\rotatebox{90}{\textbf{QA}}} & \parbox{15cm}{
\textbf{Context:} \textit{Jozef Murgaš sa narodil v Tajove. Bol synom Jána Murgaša a Zuzany Murgašovej (rod. Slamovej). Základnú školu absolvoval v rodnom Tajove, neskôr študoval na gymnáziu v Banskej Bystrici (1876 – 1880), ale zaujímalo ho predovšetkým maliarstvo. V r. 1880 – 1882 študoval v bratislavskom seminári a neskôr do roku 1884 v ostrihomskom. ...} \\
\textbf{Question:} \textit{Na akej škole študoval Jozef Murgaš v Banskej Bystrici ?} \\
\underline{\textbf{Answer:}} \underline{na gymnáziu}} \\

\bottomrule
    
\end{tabular}
\caption{Samples from the development sets of skLEP. The inputs in each sample are \textbf{bolded}, the answer category \underline{\textbf{bolded and underlined}} and the actual expected outputs (labels) are \underline{underlined}. Translation can be found in Table~\ref{tab:sklep-sample-translations}.}
\label{tab:sklep-sample}
\end{table*}

\section{Dataset Construction and Preprocessing}

Our objective in constructing the benchmark was to develop a principled tool for evaluating the language understanding capabilities of the tested models. Since most tasks were introduced in previous work, we maintained the original setup wherever possible to enable direct comparisons.

Some tasks (HS and QA) did not include a validation (development) split. We addressed this by sampling from their training sets to create a validation set matching the size of the test set. We also observed that the XNLI and STS datasets contained duplicates, which were removed during preprocessing. In the STS dataset, in particular, we eliminated sentence pairs with identical textual representations that were assigned a score other than 5—likely an artifact of the translation pipeline described below.

\section{Tasks}
The skLEP benchmark comprises three task types: token classification, sentence-pair
tasks, and document classification. Below, we describe the datasets, their tasks,
creation processes, and any major modifications made for inclusion in skLEP.

\subsection{Token-Level Tasks}

\paragraph{Universal Dependencies (UD)}
The Universal Dependencies project \cite{nivre-etal-2020-universal} is a community-
driven effort to build expanding treebanks for over 100 languages using a unified
annotation scheme. It provides POS tags, lemmas, syntactic dependencies, arguments, and
modifiers in more than 200 treebanks. For skLEP, we use the POS subset from the Slovak
Dependency Treebank \cite{11234/1-1822} in UD.

\paragraph{Universal NER (UNER)}
The Universal NER project \cite{mayhew-etal-2024-universal} offers high-quality,
cross-lingually consistent annotations for multilingual Named Entity Recognition.
Using the same data as UD, its Slovak subset (from the Slovak Dependency Treebank
\cite{11234/1-1822}) includes human-annotated labels for persons (PER), organizations
(ORG), and locations (LOC).

\paragraph{WikiGoldSK (WGSK)}
WikiGoldSK \cite{suba-etal-2023-wikigoldsk} is a manually annotated NER dataset for
Slovak that addresses the limitations of silver-standard resources. Sourced from Slovak
Wikipedia and annotated following guidelines inspired by the BSNLP-2017 Shared Task
\cite{piskorski2017first}, it uses the CoNLL-2003 NER tagset and adds a miscellaneous
(MISC) category for entities such as movies, awards, events, and media outlets.

\subsection{Sentence-Pair Tasks}

\paragraph{Recognizing Textual Entailment (RTE)}
Originating from the GLUE benchmark \cite{wang2018glue}, the RTE task merges datasets
from various entailment challenges \cite{dagan2005pascal,bar2006second,
giampiccolo2007third,bentivogli2009fifth} into a binary classification problem by
combining neutral and contradiction into a single ``not entailment'' class. As no Slovak
version existed, we translated the English dataset and post-edited it with a native speaker.
Since the original test labels were unavailable, we manually re-labeled a subset of test set translated to Slovak,
which means the final test differ from its English counterpart (1660 Slovak vs. 3000 English labeled samples).

\paragraph{Natural Language Inference (NLI)}
NLI assesses a system's ability to determine the inferential relationship between two
sentences: entailment, contradiction, or neutrality. While GLUE uses the MNLI dataset
\cite{williams2017broad}, we adopt XNLI \cite{conneau2018xnli}. As Slovak was not
pre-translated, we applied our standard translation pipeline with native speaker post-editing.

\paragraph{Semantic Textual Similarity Benchmark (STS)}
The STS task measures the semantic similarity of two sentences on a scale from 0 (none) to
5 (exact match). The original labels \cite{cer2017semeval} were averaged from multiple
annotations. In the absence of a Slovak STS dataset, we used our translation pipeline with
native post-editing while retaining the original golden labels.

\subsection{Document-Level Tasks}

\paragraph{Hate Speech Classification (HS)}
This dataset is from the Slovak Hate Speech and Offensive Language Database, designed
to detect hateful and offensive social media posts. Public posts are labeled binary (1 for
hateful/offensive, 0 otherwise). Raw posts were scraped and cleaned via text clustering to
remove spam. Experts annotated the refined posts, filtering out annotators with overly uniform
responses (over 90\% identical) or below 70\% agreement. For posts with multiple annotations,
votes from reliable annotators were aggregated, discarding cases dominated by a neutral vote.
Originally provided with only train and test sets, a validation set was created by splitting
the training data.

\paragraph{Sentiment Analysis (SA)}
Originally introduced as \textbf{Reviews3} \cite{pecar-etal-2019-improving} and later
reprocessed in \cite{gurgurov2025gremlinrepositorygreenbaseline}, the SA dataset contains
Slovak customer reviews manually labeled as positive, negative, or neutral by two annotators
reaching consensus. For skLEP, we use the version from
\cite{gurgurov2025gremlinrepositorygreenbaseline}, which merges the three classes into two by
excluding the neutral category and introducing a new split.

\paragraph{Question Answering (SK-QuAD)}
SK-QuAD \cite{10082887} is the first manually annotated Slovak question-answering dataset,
aligned with SQuAD v2.0. It contains over 91,000 Q\&A pairs, including both answerable and
unanswerable items. Developed from Slovak Wikipedia across 14,063 categories, the dataset was
created by over 150 volunteers and 9 part-time annotators, then validated by five paid reviewers.
Annotations were performed using Prodigy with preprocessing and typo detection. Data was filtered
to remove entries with severe grammatical issues or vagueness, and answers were categorized by
type. Since the original release had only train and test sets, a validation set was generated by
splitting the training data.

\begin{table}[t]
    \centering
    \begin{tabular}{lrrr}
    \toprule
    \textbf{Model} & mean & median \\
    \midrule
    DeepL & 1.81 (1.26) & 1 \\
    GPT-4o & 1.85 (1.21) & 1 \\
    Google Translate & 2.05 (1.45) & 1 \\
    MADLAD-400-3B & 2.54 (1.60) & 2 \\
    NLLB-3.3B & 2.68 (1.61) & 3 \\
    \bottomrule
    \end{tabular}
    \caption{Translation quality ranking results. The mean column represents the average rank (with standard deviation in parentheses), and the median column shows the median rank for each model.}
    \label{tab:translation-results-rank}
\end{table}

\section{Machine Translation Quality Assessment}

Due to the limited availability of high-quality Slovak datasets for general language understanding, we translated established English tasks into Slovak to construct a robust benchmark. To ensure methodological rigor, we conducted a Slovak machine translation experiment using 90 sentences, equally drawn from the original NLI, STS, and RTE datasets. 

The experiment employed five translation systems: Google Translate, DeepL, GPT-4o (prompted with “Translate the given text to Slovak”), MADLAD-400-3B \cite{kudugunta2024madlad}, and NLLB-3.3B \cite{costa2022no}. The resulting translations were evaluated by four native annotators (also co-authors of this work) following the methodology outlined in \cite{briva2024large}.

Translation quality was assessed using two approaches. First, annotators ranked the translations from each model/service in descending order; in the event of a tie, they skipped to the next available rank. Second, annotators graded each translation on two dimensions—fluency (i.e., the extent to which the translation is coherent in the target language) and adequacy (i.e., the extent to which the translation conveys the original meaning)—using a 4-point Likert scale. Prior to annotation, annotators were instructed to review and adhere to custom Translation Quality Evaluation guidelines adopted from \cite{briva2023impact} (see Appendix \ref{apx:guidelines}).

Table~\ref{tab:translation-results-rank} presents the results of the ranking experiment. The top-performing systems—DeepL, GPT-4o, and Google Translate—consistently achieved lower (i.e., better) average and median rankings, indicating superior translation quality compared to MADLAD-400-3B and NLLB-3.3B.

Table~\ref{tab:translation-results-fluency-adequacy} shows the results of the fluency/adequacy evaluation. DeepL achieved the highest fluency (3.70 ± 0.53) and adequacy (3.73 ± 0.51) scores. GPT-4o matched DeepL in adequacy (3.73 ± 0.54) but scored slightly lower in fluency (3.62 ± 0.59). Google Translate received marginally lower ratings, while both MADLAD-400-3B and NLLB-3.3B demonstrated significantly lower performance, with NLLB-3.3B obtaining the lowest fluency score (3.40 ± 0.75).

\begin{table}[t]
    \centering
    \begin{tabular}{lrr}
    \toprule
    \textbf{Model} & \textbf{Fluency} & \textbf{Adequacy} \\
    \midrule
    DeepL & 3.70 (0.53) & 3.73 (0.51) \\
    GPT-4o & 3.62 (0.59) & 3.73 (0.54) \\
    Google Translate & 3.57 (0.65) & 3.67 (0.62) \\
    MADLAD-400-3B & 3.48 (0.75) & 3.53 (0.73) \\
    NLLB-3.3B  & 3.40 (0.75) & 3.54 (0.71) \\
    \bottomrule
    \end{tabular}
    \caption{Fluency and adequacy evaluation results. The reported values represent the mean scores with the corresponding standard deviations in parentheses.}
    \label{tab:translation-results-fluency-adequacy}
\end{table}

Based on these results, we selected DeepL for translation. However, due to its commercial cost for large-scale translation, we employed MADLAD-400-3B to translate the NLI corpus, which is an order of magnitude larger than the other datasets and would be prohibitively expensive to process with a commercial service. This has allowed us to keep the translation costs manageable, on the order of 100 EUR for all of the datasets.

To further ensure the high quality of the translation (and post-editing) pipeline we utilize, we conducted two additional experiments, which are described in more detail in Appendix \ref{sec:relabel} and Appendix \ref{sec:post-editing}.

\section{Experiments}

We use the proposed benchmark to evaluate various pre-trained language models (PLMs) based on the encoder-only Transformer architecture. Our selection includes well-established multilingual models, Slovak-specific models, and monolingual English variants, as the latter have been shown to transfer well to non-English languages \cite{artetxe2019cross,blevins2022language}. This evaluation aims to assess the impact of different model variants, sizes, and pre-training regimes on overall performance.

We provide a reference implementation for each of the considered tasks using the HuggingFace Transformers framework \cite{wolf-etal-2020-transformers} and conduct an extensive hyperparameter grid search across all tasks, models and various hyperparameter settings by finetuning the models on the specific task's training data and evaluating it on its development split.  More details can be found in Appendix \ref{sec:hyper}.

\subsection{Slovak-specific models}

\paragraph{SlovakBERT} Introduced in
\cite{pikuliak2021slovakbert}, SlovakBERT is a base-sized masked language model built on the
RoBERTa architecture and tailored specifically for Slovak. It is the first model designed solely for
Slovak natural language processing tasks. Trained on a 19.35GB web-crawled corpus, it outperforms
multilingual models in both efficiency and accuracy by using a vocabulary and training data
exclusively curated for Slovak.

\paragraph{HPLT BERT$_{base-sk}$} HPLT BERT$_{base-sk}$ is a Slovak-specific masked language model
derived from LTG-BERT \cite{samuel-etal-2023-trained}, an optimized variant of the classic BERT
architecture. It is part of the HPLT project’s series of monolingual, encoder-only models
\cite{de-gibert-etal-2024-new-massive}, which were trained across 75 languages using language-specific
tokenizers and datasets. The Slovak subset of the training data, drawn from the HPLT project's 1.2 data
release, amounts to 33.4GB.

\paragraph{FERNET-CC\_sk} FERNET-CC\_sk, introduced in \cite{FERNETC5}, is a monolingual Slovak
BERT-base model pre-trained on 29GB of filtered Slovak Common Crawl data. It adheres to the standard
BERT-base architecture with 12 layers, 12 attention heads, and a hidden size of 768, and employs a
SentencePiece tokenizer with a 100K-token vocabulary specifically designed for Slovak orthography.

\subsection{English-specific models}

\paragraph{DeBERTaV3} Introduced in \cite{he2021debertav3}, DeBERTaV3 is an enhanced version of the
DeBERTa language model. It integrates ELECTRA-style Replaced Token Detection (RTD) for more efficient
pre-training. Built on a Transformer-based architecture, it replaces masked language modeling (MLM)
with RTD, wherein a generator produces token replacements and a discriminator determines whether each
token is original or replaced. To address training inefficiencies, DeBERTaV3 employs Gradient-
Disentangled Embedding Sharing (GDES), which prevents conflicting gradient updates between the generator
and discriminator, thereby optimizing representation learning.

\paragraph{ModernBERT} ModernBERT, introduced in
\cite{warner2024smarterbetterfasterlonger}, is a cutting-edge encoder-only transformer model engineered
for superior downstream performance and efficiency, particularly with long sequences. It supports a native
context length of 8,192 tokens and incorporates several architectural innovations, including GeGLU
activation, RoPE positional embeddings, and an alternating local-global attention mechanism. Trained on
2 trillion tokens of primarily English data, ModernBERT is available in two variants: ModernBERT$_{Base}$
(with 149M parameters and 22 layers) and ModernBERT$_{Large}$ (with 395M parameters and 28 layers).

\begin{table*}[t]
\centering
\resizebox{\textwidth}{!}{%
\begin{tabular}{l || c | c || c | c || c | c || c | c || c | c || c | c || c | c || c | c || c | c || c | c ||}
\toprule
\multirow{3}{*}{\textbf{Model}} & 
\multicolumn{2}{c}{} &
\multicolumn{6}{c}{\textit{Token-Level}} & 
\multicolumn{6}{c}{\textit{Sentence-Pair}} & 
\multicolumn{6}{c}{\textit{Document-Level}} \\
\cmidrule(lr){4-9}\cmidrule(lr){10-15}\cmidrule(lr){16-21}
& 
\multicolumn{2}{c}{\textbf{AVG}} & 
\multicolumn{2}{c}{\textbf{UD}} & 
\multicolumn{2}{c}{\textbf{UNER}} & 
\multicolumn{2}{c}{\textbf{WGSK}} & 
\multicolumn{2}{c}{\textbf{RTE}} & 
\multicolumn{2}{c}{\textbf{NLI}} & 
\multicolumn{2}{c}{\textbf{STS}} & 
\multicolumn{2}{c}{\textbf{HS}} & 
\multicolumn{2}{c}{\textbf{SA}} & 
\multicolumn{2}{c}{\textbf{QA}} \\
& RER & Avg. &
RER & F1 & 
RER & F1 & 
RER & F1 & 
RER & Acc. & 
RER & Acc. & 
RER & Paer. & 
RER & Acc. & 
RER & Acc. & 
RER & F1 \\

\midrule
\includegraphics[width=0.3cm]{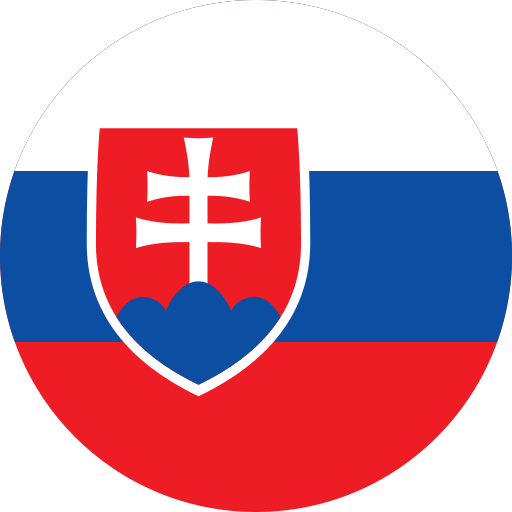}
SlovakBERT & 0.0 &  83.95
& 0.00 & 98.04
& 0.00 & 81.92
& 0.00 & 92.14
& 0.00 & 65.20
& 0.00 & 82.75
& 0.00 & 83.18
& 0.00 & 80.31
& 0.00 & 97.63
& 0.00 & 74.36
\\

\midrule

\includegraphics[width=0.3cm]{latex/images/sk.png}

HPLT\textsubscript{Base} &  -1.29 & 82.96
&  9.71 & 98.23
& 13.05 & 84.28
& 16.50 & 93.44
& -24.12 & 56.81
& -11.83 & 80.71
& -9.01 & 81.66
& -7.68 & 78.80
& -1.23 & 97.60
& 3.04 & 75.14
\\

\includegraphics[width=0.3cm]{latex/images/sk.png}
FERNET-CC\textsubscript{Base} &  0.55 & 84.27
& -8.67 & 97.87
& 9.22 & 83.59
& 8.12 & 92.78
& 8.72 & 68.23
& -5.31 & 81.83
& 2.50 & 83.60
& -6.40 & 79.05
& -1.23 & 97.60
& -1.97 & 73.85	
\\

\midrule

\includegraphics[width=0.3cm]{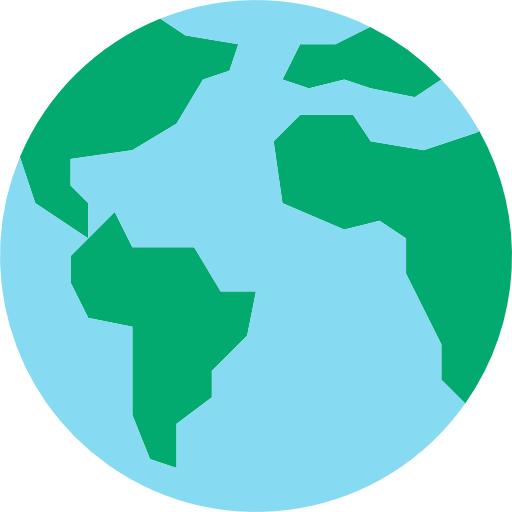}
DistilmBERT\textsubscript{Base} & -41.42 & 78.54
& -45.84 & 97.14
& -34.96 & 75.60
& -59.00 & 87.50
& -16.79 & 59.36
& -50.45 & 74.05
& -56.38 & 73.70
& -24.24 & 75.54
& -59.27 & 96.23
& -25.84 & 67.73
\\

\includegraphics[width=0.3cm]{latex/images/world.png}
MiniLM\textsubscript{L12-Base} &  -41.21 & 78.75
& -36.16 & 97.33
& -67.55 & 69.71
& -82.88 & 85.63
& -18.75 & 58.67
& -52.50 & 73.69
& -23.25 & 79.27
& -23.21 & 75.74
& -59.27 & 96.23
& -7.35 & 72.47
\\

\midrule

\includegraphics[width=0.3cm]{latex/images/world.png}
mBERT\textsubscript{Base} &	-30.82 & 81.16
& -78.38 & 96.50
& -32.51 & 76.04
& -56.79 & 87.68
& 2.83 & 66.18
& -28.09 & 77.90
& -26.21 & 78.77
& -15.00 & 77.36
& -39.03 & 96.71
& -4.23 & 73.28
\\

\includegraphics[width=0.3cm]{latex/images/world.png}
XLM-R\textsubscript{Base} & -7.58 & 82.30
&  -6.09 & 97.92
& -1.29 & 81.69
& -2.05 & 91.98
& -17.42 & 59.14
& -9.36 & 81.14
& -14.59 & 80.73
& -16.03 & 77.15
& 2.82 & 97.70
& -4.20 & 73.28
\\

\includegraphics[width=0.3cm]{latex/images/world.png}
XLM-V\textsubscript{Base} & -33.83 & 80.56
& -20.69 & 97.63
& -28.86 & 76.70
& -14.13 & 91.03
& -21.98 & 57.55
& -15.62 & 80.06
& -19.84 & 79.84
& -29.50 & 74.50
& -151.06 & 94.05
& -2.76 & 73.65
\\

\includegraphics[width=0.3cm]{latex/images/world.png}
mDeBERTaV3\textsubscript{Base} &  \textbf{6.43} & \textbf{85.17}
& -1.22 & 98.02
& 8.40 & 83.44
& 9.51 & 92.89
& 16.51 & 70.94
& 9.60 & 84.41
& 9.76 & 84.82
& -10.25 & 78.29
& 9.56 & 97.86
& 5.96 & 75.89
\\

\midrule

\includegraphics[width=0.3cm]{latex/images/world.png}
XLM-R\textsubscript{Large} &  -11.90 & 83.36
&  9.95 & 98.23
& 7.19 & 83.22
& 1.29 & 92.24
& -22.56 & 57.35
& 17.71 & 85.80
& 24.80 & 87.35
& -30.91 & 74.22
& -125.41 & 94.66
& 10.86 & 77.14
\\

\midrule

\includegraphics[width=0.3cm]{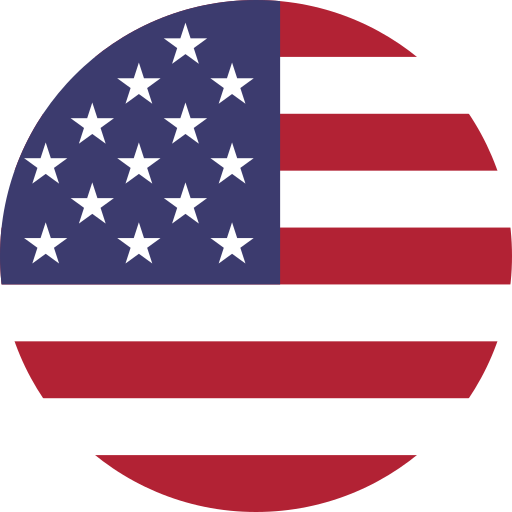}
DeBERTaV3\textsubscript{Base} & -52.40 & 78.48
& -77.17 & 96.53
& -66.97 & 69.81
& -71.13 & 86.55
& -10.73 & 61.47
& -27.20 & 78.06
& -48.12 & 75.09
& -38.62 & 72.71
& -120.01 & 94.79
& -11.65 & 71.37
\\

\includegraphics[width=0.3cm]{latex/images/us.png}
DeBERTaV3\textsubscript{Large} &  -10.22 & 83.95
& -25.22 & 97.55
& -18.13 & 78.64
& -22.60 & 90.36
& 26.14 & 74.30
& 13.77 & 85.13
& -7.38 & 81.94
& -21.03 & 76.17
& -39.03 & 96.71
& 1.45 & 74.73	
\\

\includegraphics[width=0.3cm]{latex/images/us.png}
ModernBERT\textsubscript{Base} &  -132.43 & 72.82
& -332.62 & 91.52
& -184.06 & 48.64
& -282.12 & 69.97
& -21.35 & 57.77
& -65.66 & 71.42
& -57.07 & 73.58
& -31.68 & 74.07
& -84.92 & 95.62
& - & -
\\

\includegraphics[width=0.3cm]{latex/images/us.png}
ModernBERT\textsubscript{Large} & -110.09 & 74.46
& -212.22 & 93.88
& -169.39 & 51.29
& -171.84 & 78.63
& -36.23 & 52.59
& -43.85 & 75.19
& -43.06 & 75.94
& -28.73 & 74.65
& -175.35 & 93.47
& - & -
\\

\bottomrule
\end{tabular}
    } 
\caption{Task-specific scores and Relative Error Reduction (RER) are reported for each evaluated model. Scores were obtained by fine-tuning every model on the corresponding task’s training set with three seeds (12, 42 and 99) and evaluating on the test set; the reported value is the mean across these runs. Models are grouped by the dominant language of their training corpora and by model size. The best average absolute performance score and RER is bolded.}

\label{tab:results}
\end{table*}

\subsection{Multilingual models}

\paragraph{mBERT$_{Base}$} Introduced in \cite{devlin-etal-2019-bert}, mBERT$_{Base}$ is the first
multilingual BERT model. Pre-trained on 104 Wikipedias (including Slovak) using a self-supervised
strategy that combines Masked Language Modeling (MLM) and Next Sentence Prediction (NSP), it employs
WordPiece tokenization with a shared vocabulary of 110,000 tokens. The model also incorporates
language-specific preprocessing to handle diverse scripts. By masking 15\% of tokens during pre-training
and predicting sentence continuity, it laid the foundation for effective cross-lingual transfer in downstream
tasks.

\paragraph{DistilmBERT} DistilmBERT is a distilled variant of the BERT base multilingual cased model
presented in \cite{Sanh2019DistilBERTAD} and pre-trained on the same dataset as mBERT$_{Base}$. Featuring 6
layers, a 768-dimensional hidden size, 12 attention heads, and 134M parameters, it is approximately 25\%
smaller and nearly twice as fast as its 177M-parameter counterpart. The distillation process retains
essential multilingual features while providing improved computational efficiency for downstream tasks.

\paragraph{XLM-R} Introduced in \cite{conneau-etal-2020-unsupervised}, XLM-R is a large-scale
multilingual masked language model designed for cross-lingual representation learning. Based on a
Transformer architecture, it was pre-trained on over two terabytes of filtered CommonCrawl data covering
100 languages, including 23.2GB of Slovak. In contrast to earlier models like mBERT and XLM, XLM-R uses an
expanded 250K-token vocabulary trained via SentencePiece tokenization, enhancing efficiency across diverse
linguistic structures. It employs a multilingual masked language modeling objective and incorporates
optimizations in language sampling, model capacity, and vocabulary allocation to alleviate the
``curse of multilinguality.'' The model is available in two variants: XLM-R Base (12 layers, 768 hidden size,
270M parameters) and XLM-R Large (24 layers, 1024 hidden size, 550M parameters).

\paragraph{XLM-V} Introduced in \cite{liang-etal-2023-xlm}, XLM-V is a multilingual masked language model
aimed at addressing the vocabulary bottleneck in large-scale multilingual NLP. Trained on the same dataset
as XLM-R, it utilizes an expanded vocabulary of one million tokens, reducing the need for token sharing between
linguistically distant languages. Compared to XLM-R, XLM-V generates more semantically meaningful and compact
tokenizations, thereby enhancing language representation across diverse linguistic groups—a benefit that has
proven especially effective for downstream tasks in low-resource languages.

\paragraph{mDeBERTaV3} Introduced in \cite{he2021debertav3}, mDeBERTaV3 extends DeBERTaV3 to the
multilingual setting. It was trained on the 2.5TB CC100 dataset---the same dataset used for XLM-R---but with only
one-third as many training passes. The model retains Disentangled Attention (DA) and Replaced Token Detection
(RTD), while also employing Gradient-Disentangled Embedding Sharing (GDES) to optimize multilingual token
representations. Using a 250K-token SentencePiece vocabulary, mDeBERTaV3 achieves state-of-the-art
cross-lingual performance, surpassing XLM-R in zero-shot transfer on benchmarks such as XNLI.

\subsection{Evaluation Metrics}
We report task-specific metrics, such as the F1-score for classification tasks, accuracy for QA and NLI, and Pearson correlation for regression-based tasks. When aggregating scores into a single overall metric, existing benchmarks typically compute a simple average across tasks. This approach, however, assumes that absolute score differences carry equal weight across all tasks, which is a strong assumption.

While this assumption may hold for a homogeneous task set, our benchmark encompasses tasks with significant variations in absolute score ranges. For example, the F1 score for the UD task is expected to exceed 95, whereas the QA task is likely to have an F1 score around 75. Consequently, a one-point change in the UD task, which corresponds to a 20\% error reduction, is far more impactful than the same change in the QA task, where it only reduces error by approximately 3\%.

To address this disparity, we adopt the approach introduced in \cite{de2023dumb} and report Relative Error Reduction (RER) compared to a baseline. We use SlovakBERT as our baseline and provide both the average RER and the average score across all skLEP task metrics.

\subsection{Results}

 While the results of the hyperparameter search can be found in Appendix \ref{sec:hyper}, Table~\ref{tab:results} present the main results. Overall, we can conclude that despite its age, SlovakBERT can still be considered a potent baseline, as only two models reported higher mean absolute scores on the skLEP benchmark, with mDeBERTaV3\textsubscript{Base} scoring the highest while also obtaining a Relative Error Reduction (RER) of 6.43 percentage points. Although RER generally tends to mimic the trend of the mean absolute scores, we can see its usefulness in the case of DeBERTaV3\textsubscript{Base} which reported the exact same mean absolute score as SlovakBERT, but at the same time a negative RER as well, which suggests that SlovakBERT may be a better choice where a balanced performance across various tasks in Slovak is desired. A closer look at the absolute scores of the models across the respective tasks suggests that while some of them might be considered close to being solved (e.g. \textbf{UD} or \textbf{SA}, for which the majority of models report F1 score and accuracy of over 90.0), there is substantial room for improvement, particularly for the \textbf{QA} and \textbf{RTE} tasks, whose F1 score and accuracy is generally reported to be below 75.0 and 70.0, respectively.

\section{Discussion}

\paragraph{Software Tools}
We developed the skLEP benchmark as an open-source toolkit that can help researchers and
practitioners choose models for specific tasks and supports the development and evaluation
of new models. Built on the established HuggingFace Transformers framework
\cite{wolf-etal-2020-transformers}, all datasets are prepared for integration with the
HuggingFace Datasets repository \cite{lhoest-etal-2021-datasets}. We will release all software
and data upon acceptance.

\paragraph{Dataset Licenses}
To ensure continuity, the original dataset licenses remain unchanged and are summarized in
Table~\ref{tab:dataset-licensing}, along with additional reference details. All datasets are
sourced from public repositories and are available for academic use. Although the translated
datasets are not yet public, we plan to release them upon publication.

\begin{table}[ht]
\centering

\resizebox{0.45\textwidth}{!}{%

\begin{tabular}{lcccc}
\toprule
\textbf{Task} & \textbf{Translated} & \textbf{License} & \textbf{Website} & \textbf{Code} \\ \midrule
UD         &         & \faCreativeCommons         & \href{https://universaldependencies.org/}{\faLink}          &  \href{https://github.com/UniversalDependencies/UD\_Slovak-SNK/}{\faCode}             \\ 
UNER       &         &  \faCreativeCommons & \href{https://universalner.org/}{\faLink}          &  \href{https://github.com/UniversalNER/UNER\_Slovak-SNK/}{\faCode}               \\
WGSK       &         & \faCreativeCommons &           & \href{https://github.com/NaiveNeuron/WikiGoldSK}{\faCode}       \\ 
RTE       & \faLanguage        & \faTimes & \href{https://aclweb.org/aclwiki/Recognizing\_Textual\_Entailment}{\faLink}                &              \\ 
NLI          & \faLanguage        & \faGraduationCap & \href{https://huggingface.co/datasets/facebook/xnli}{\faLink}                & \href{https://github.com/facebookresearch/XNLI}{\faCode}       \\ 
STS         & \faLanguage        & \faTimes  & \href{https://web.archive.org/web/20240319092902/https://ixa2.si.ehu.eus/stswiki/index.php/STSbenchmark}{\faLink}              & \href{https://github.com/PhilipMay/stsb-multi-mt/}{\faCode}       \\
HS        &         & \faCreativeCommons      & \href{https://huggingface.co/datasets/TUKE-KEMT/hate_speech_slovak}{\faLink}                &        \\ 
SA         &         & \faLockOpen & \href{https://huggingface.co/datasets/DGurgurov/slovak_sa}{\faLink}          & \href{https://github.com/pyRis/retrofitting-embeddings-lrls}{\faCode}          \\ 
QA         &         & \faCreativeCommons         &   \href{https://huggingface.co/datasets/TUKE-DeutscheTelekom/skquad}{\faLink}               &      \\ 
\bottomrule
\end{tabular}
}
\caption{Summary information on the skLEP datasets, their translation status, license, as well as references. The icons denote whether the dataset was \faLanguage~ translated, whether it is released under the terms of \faCreativeCommons~ creative commons license allowing commercial use, \faGraduationCap~ non-commercial use only, \faLockOpen~ MIT license or \faTimes~ no specific license; whether the authors provide \faLink~ dataset website and a \faCode~ code repository.}
\label{tab:dataset-licensing}
\end{table}

\paragraph{Leaderboard}
All evaluation results are compiled into the skLEP leaderboard, offering a standardized way
to assess the current state-of-the-art for various Slovak NLU tasks. The leaderboard links model
metadata (e.g. size as shown in Table~\ref{tab:model_params}), configuration details, performance on
both dev and test splits, and the scripts used for evaluation, following the guidelines in
\cite{ethayarajh-jurafsky-2020-utility}.

This approach enhances transparency but has drawbacks. With public test sets, there is a risk of
their inadvertent use in future training, and it deviates from the original (Super)GLUE framework
\cite{wang2018glue,wang2019superglue}. However, as most skLEP datasets were already public,
we opted for greater transparency. This also eliminates the need for benchmark authors to be involved
in evaluation—a method adopted by recent work \cite{berdivcevskis2023superlim}.

Finally, since leaderboards tend to saturate over time, we plan to open our platform to researchers
to incorporate their datasets, design new benchmarks \cite{ma2021dynaboard,thrush-etal-2022-dynatask}, and
develop additional Slovak resources using a human-and-model-in-the-loop approach
\cite{kiela-etal-2021-dynabench}.

We intend to unveil the leaderboard and all associated data upon acceptance.

\section{Related Work}

The release of the GLUE \cite{wang2018glue} and SuperGLUE \cite{wang2019superglue} benchmarks has made it much easier to evaluate pretrained language models on English NLU tasks. In the case of GLUE, comprised of nine tasks in total, these were dominated by four Natural Language Inference (NLI) tasks. On the other hand, half of the more sophisticated SuperGLUE's tasks are focused on Question Answering (QA). 

The English-specific GLUE and SuperGLUE have also inspired the creation of similar benchmarks for various other languages: CLUE~\cite{xu-etal-2020-clue} in Chinese , KLEJ~\cite{rybak2020klej} and LEPISZCZE~\cite{augustyniak2022way} in Polish, CLUB in Catalan \cite{rodriguez2021catalan},  IndoNLU for Indonesian \cite{wilie2020indonlu},  LiRo in Romanian \cite{dumitrescu2021liro}, ParsiNLU in Persian \cite{khashabi2021parsinlu},  Slovene SuperGLUE in Slovenian \cite{vzagar2022slovene}, NorBench in Norwegian \cite{samuel2023norbench}, KLUE in Korean \cite{park2021klue}, FLUE in French \cite{le2019flaubert}, JGLUE for Japanese \cite{kurihara2022jglue}, ORCA in Arabic \cite{elmadany2022orca}, BasqueGLUE in Basque \cite{urbizu2022basqueglue}, RussianSuperGLUE in Russian \cite{shavrina2020russiansuperglue}, DUMB in Dutch \cite{de2023dumb}, SuperGLEBer in German \cite{pfister2024supergleber}, UINAUIL~\cite{basile2023uinauil} and Invalsi~\cite{puccetti-etal-2025-invalsi} in Italian, PORTULAN ExtraGLUE in Portuguese \cite{osorio2024portulan} and Superlim in Swedish \cite{berdivcevskis2023superlim}. They generally assess core language understanding through tasks such as natural language inference, question answering, sentiment analysis, and named entity recognition (NER), along with syntactic and semantic evaluations (e.g., POS tagging, dependency parsing, and semantic similarity) and various language-specific diagnostic or classification challenges.


To evaluate models across multiple languages, various multilingual benchmarks have also been proposed. The most prominent ones include XGLUE \cite{liang2020xglue}, XTREME \cite{hu2020xtreme} and XTREME-R \cite{ruder2021xtreme} covering 19, 40 and 50 languages, respectively. Unfortunately, none of these benchmarks cover the Slovak language. Moreover, they only provide English training data, making them more suitable for evaluating language transfer across languages rather than language-specific performance.

The work conceptually and in spirit closest to ours would be the introduction of the SlovakBERT model \cite{pikuliak2021slovakbert}. In it, the authors evaluate it against a set of other models on part-of-speech tagging, semantic textual similarity, sentiment analysis
and document classification. Of these, the part-of-speech dataset is the same one skLEP uses, the semantic textual similarity task has been automatically translated using an machine translation model which has since been surpassed by the ones we evaluate in our experiments, and the document classification task can be considered solved, as SlovakBERT reports Macro-F1 score of 99 out of 100. In skLEP we build upon this work and extend it into a fully featured Slovak general language understanding benchmark. 

\section{Conclusion}
skLEP introduces the first comprehensive Slovak NLU benchmark, addressing the lack of standardized evaluation resources for the language. By curating and translating datasets, we provide a diverse suite of nine tasks spanning token-level, sentence-pair, and document-level challenges. Our evaluation of Slovak-specific, multilingual, and English models highlights the strengths and limitations of existing approaches. The results indicate that, while Slovak-specific transformers remain competitive, the largest error reduction is delivered by parameter-efficient multilingual DeBERTa variants, suggesting cost-aware directions for future model design. We publicly release the benchmark, models, and evaluation toolkit to foster transparency and further advancements. Future work includes expanding skLEP with additional tasks and improving dataset quality through human annotation.

\section*{Limitations}

\subsection*{Tasks Included in skLEP} 
The proposed benchmark comprises nine tasks, evenly distributed among three categories: Token-Level, Sentence-Pair, and Document-Level. Some tasks within each category are relatively similar; for example, we include two Named Entity Recognition (NER) datasets (albeit with different tag sets), and the Natural Language Inference (NLI) and Recognizing Textual Entailment (RTE) tasks both address similar concepts. However, in our benchmark, NLI is formulated as a three-way classification task, whereas RTE is a two-way classification. Although incorporating additional tasks would broaden the coverage of natural language understanding, we are constrained by the availability of high-quality Slovak datasets. Consequently, we could not include other NLP tasks that are considered standard for higher-resource languages, nor tasks involving non-textual or multimodal data. We hope that this release will encourage the community to contribute further high-quality tasks to the benchmark, which would be greatly appreciated.

\subsection*{Translation} 
Three tasks in the benchmark—namely, RTE, NLI, and STS—were automatically translated, and their training sets have not been manually corrected. Although the validation and test sets were manually corrected, they may still exhibit characteristics of \emph{translationese} \cite{koppel-ordan-2011-translationese} and its more pronounced form, \emph{post-editese} \cite{toral-2019-post}. We mitigated these issues by employing native speakers with backgrounds in NLP and/or linguistics for post-editing; however, some artifacts may persist.

\subsection*{Evaluation}
We exclusively evaluate Transformer-based encoder-only models and have structured the benchmark primarily for fine-tuning these models, making it less suitable for generative models. Our goal with the initial release of the skLEP benchmark is to establish robust baselines through a standard fine-tuning approach combined with an extensive hyperparameter search. This foundation is intended to better contextualize future work on generative or prompt-engineering based solutions, which often exhibit higher variability.

The Slovak-specific models evaluated in this work are all of the “base” size (defined as having fewer than 150 million parameters in \cite{warner2024smarterbetterfasterlonger}), and our evaluation includes a greater number of English models than Slovak ones. This imbalance is due to the limited availability of Slovak models, and we hope that the release of this benchmark will inspire further development in this area.

Our hyperparameter search and model selection were constrained by our computational budget, which is particularly visible in the case of the NLI dataset, whose large size prevented us from exploring further hyperparameters. Despite the extensive nature of our search, it is conceivable that better-performing hyperparameters or models exist. With this initial release, we encourage the community to contribute additional models and configurations to the benchmark; we also plan to extend this work in the future.

Currently, the skLEP benchmark does not include a human baseline, which may make it challenging to contextualize the performance of the evaluated models.

\section*{Ethics Statement}

\subsection*{Dataset}
The benchmark comprises only publicly available datasets and derivatives thereof, each licensed to permit free use for academic research—most under Creative Commons licenses. Wherever applicable, we have cited the original works that introduced the respective tasks and datasets. Users of skLEP are encouraged to consult these sources for detailed licensing information.

For some datasets, we created new splits, removed duplicates, or added our own annotations and re-annotations. We intend to release both the updated datasets and the scripts used in their creation upon acceptance.

\subsection*{Intended Use}
The skLEP benchmark is intended to foster the evaluation and development of language models specifically for the Slovak language. To this end, we are making the code, datasets, models, configurations, and leaderboard publicly available. We encourage the community to submit additional models and tasks, and such contributions would be greatly appreciated.

\subsection*{Environmental Impact}

Although we propose the skLEP benchmark with the aim of stimulating the development of new models for Slovak, it is important to note that training such models can require substantial computational resources, potentially contributing to global warming \cite{strubell2020energy}. However, the current form of the skLEP benchmark is designed for fine-tuning, which requires considerably fewer resources. Furthermore, by conducting an extensive hyperparameter search and releasing both the benchmark and the models on HuggingFace, we aim to reduce the environmental impact while providing the community with readily available models, thereby mitigating the need for expensive hyperparameter searches and fine-tuning.

\section*{Acknowledgments}

This study was funded by the Ministry of Education, Research, Development and Youth of the Slovak Republic under the project KEGA 049TUKE-4/2024 and by the Slovak Research and Development Agency under the project APVV-22-0414.

Part of the research results was obtained using the computational resources procured in the
national project National competence centre for high performance computing (project code:
311070AKF2) funded by European Regional Development Fund, EU Structural Funds Informatization
of society, Operational Program Integrated Infrastructure.

This research was partially supported by DisAI - Improving scientific excellence and creativity in combating disinformation with artificial intelligence and language technologies, a project funded by European Union under the Horizon Europe, GA No. 101079164, and by the Slovak Research and Development Agency under the Contract no. APVV-21-0114.

The flag icons have been designed using resources from Flaticon.com

\bibliography{custom,anthology}

\begin{thebibliography}{73}
\providecommand{\natexlab}[1]{#1}

\bibitem[{Artetxe et~al.(2019)Artetxe, Ruder, and Yogatama}]{artetxe2019cross}
Mikel Artetxe, Sebastian Ruder, and Dani Yogatama. 2019.
\newblock On the cross-lingual transferability of monolingual representations.
\newblock \emph{arXiv preprint arXiv:1910.11856}.

\bibitem[{Augustyniak et~al.(2022)Augustyniak, Tagowski, Sawczyn, Janiak, Bartusiak, Szymczak, Janz, Szyma{\'n}ski, W{\k{a}}troba, Morzy et~al.}]{augustyniak2022way}
Lukasz Augustyniak, Kamil Tagowski, Albert Sawczyn, Denis Janiak, Roman Bartusiak, Adrian Szymczak, Arkadiusz Janz, Piotr Szyma{\'n}ski, Marcin W{\k{a}}troba, Miko{\l}aj Morzy, et~al. 2022.
\newblock This is the way: designing and compiling lepiszcze, a comprehensive nlp benchmark for polish.
\newblock \emph{Advances in Neural Information Processing Systems}, 35:21805--21818.

\bibitem[{Bar-Haim et~al.(2006)Bar-Haim, Dagan, Dolan, Ferro, Giampiccolo, Magnini, and Szpektor}]{bar2006second}
Roy Bar-Haim, Ido Dagan, Bill Dolan, Lisa Ferro, Danilo Giampiccolo, Bernardo Magnini, and Idan Szpektor. 2006.
\newblock The second pascal recognising textual entailment challenge.
\newblock In \emph{Proceedings of the second PASCAL challenges workshop on recognising textual entailment}, volume~1. Citeseer.

\bibitem[{Basile et~al.(2023)Basile, Bioglio, Bosca, Bosco, and Patti}]{basile2023uinauil}
Valerio Basile, Livio Bioglio, Alessio Bosca, Cristina Bosco, and Viviana Patti. 2023.
\newblock Uinauil: A unified benchmark for italian natural language understanding.
\newblock In \emph{Proceedings of the 61st Annual Meeting of the Association for Computational Linguistics (Volume 3: System Demonstrations)}, pages 348--356.

\bibitem[{Bentivogli et~al.(2009)Bentivogli, Clark, Dagan, and Giampiccolo}]{bentivogli2009fifth}
Luisa Bentivogli, Peter Clark, Ido Dagan, and Danilo Giampiccolo. 2009.
\newblock The fifth pascal recognizing textual entailment challenge.
\newblock \emph{TAC}, 7(8):1.

\bibitem[{Berdi{\v{c}}evskis et~al.(2023)Berdi{\v{c}}evskis, Bouma, Kurtz, Morger, {\"O}hman, Adesam, Borin, Dann{\'e}lls, Forsberg, Isbister et~al.}]{berdivcevskis2023superlim}
Aleksandrs Berdi{\v{c}}evskis, Gerlof Bouma, Robin Kurtz, Felix Morger, Joey {\"O}hman, Yvonne Adesam, Lars Borin, Dana Dann{\'e}lls, Markus Forsberg, Tim Isbister, et~al. 2023.
\newblock Superlim: A swedish language understanding evaluation benchmark.
\newblock In \emph{Proceedings of the 2023 Conference on Empirical Methods in Natural Language Processing}, pages 8137--8153.

\bibitem[{Blevins and Zettlemoyer(2022)}]{blevins2022language}
Terra Blevins and Luke Zettlemoyer. 2022.
\newblock Language contamination helps explain the cross-lingual capabilities of english pretrained models.
\newblock \emph{arXiv preprint arXiv:2204.08110}.

\bibitem[{Briva-Iglesias et~al.(2024)Briva-Iglesias, Camargo, and Dogru}]{briva2024large}
Vicent Briva-Iglesias, Joao Lucas~Cavalheiro Camargo, and Gokhan Dogru. 2024.
\newblock Large language models" ad referendum": How good are they at machine translation in the legal domain?
\newblock \emph{arXiv preprint arXiv:2402.07681}.

\bibitem[{Briva-Iglesias et~al.(2023)Briva-Iglesias, O’Brien, and Cowan}]{briva2023impact}
Vicent Briva-Iglesias, Sharon O’Brien, and Benjamin~R Cowan. 2023.
\newblock The impact of traditional and interactive post-editing on machine translation user experience, quality, and productivity.
\newblock \emph{Translation, Cognition \& Behavior}, 6(1):60--86.

\bibitem[{Cer et~al.(2017)Cer, Diab, Agirre, Lopez-Gazpio, and Specia}]{cer2017semeval}
Daniel Cer, Mona Diab, Eneko Agirre, Inigo Lopez-Gazpio, and Lucia Specia. 2017.
\newblock Semeval-2017 task 1: Semantic textual similarity-multilingual and cross-lingual focused evaluation.
\newblock \emph{arXiv preprint arXiv:1708.00055}.

\bibitem[{Conneau et~al.(2020)Conneau, Khandelwal, Goyal, Chaudhary, Wenzek, Guzm{\'a}n, Grave, Ott, Zettlemoyer, and Stoyanov}]{conneau-etal-2020-unsupervised}
Alexis Conneau, Kartikay Khandelwal, Naman Goyal, Vishrav Chaudhary, Guillaume Wenzek, Francisco Guzm{\'a}n, Edouard Grave, Myle Ott, Luke Zettlemoyer, and Veselin Stoyanov. 2020.
\newblock \href {https://doi.org/10.18653/v1/2020.acl-main.747} {Unsupervised cross-lingual representation learning at scale}.
\newblock In \emph{Proceedings of the 58th Annual Meeting of the Association for Computational Linguistics}, pages 8440--8451, Online. Association for Computational Linguistics.

\bibitem[{Conneau et~al.(2018)Conneau, Lample, Rinott, Williams, Bowman, Schwenk, and Stoyanov}]{conneau2018xnli}
Alexis Conneau, Guillaume Lample, Ruty Rinott, Adina Williams, Samuel~R Bowman, Holger Schwenk, and Veselin Stoyanov. 2018.
\newblock Xnli: Evaluating cross-lingual sentence representations.
\newblock \emph{arXiv preprint arXiv:1809.05053}.

\bibitem[{Costa-juss{\`a} et~al.(2022)Costa-juss{\`a}, Cross, {\c{C}}elebi, Elbayad, Heafield, Heffernan, Kalbassi, Lam, Licht, Maillard et~al.}]{costa2022no}
Marta~R Costa-juss{\`a}, James Cross, Onur {\c{C}}elebi, Maha Elbayad, Kenneth Heafield, Kevin Heffernan, Elahe Kalbassi, Janice Lam, Daniel Licht, Jean Maillard, et~al. 2022.
\newblock No language left behind: Scaling human-centered machine translation.
\newblock \emph{arXiv preprint arXiv:2207.04672}.

\bibitem[{Dagan et~al.(2005)Dagan, Glickman, and Magnini}]{dagan2005pascal}
Ido Dagan, Oren Glickman, and Bernardo Magnini. 2005.
\newblock The pascal recognising textual entailment challenge.
\newblock In \emph{Machine learning challenges workshop}, pages 177--190. Springer.

\bibitem[{de~Gibert et~al.(2024)de~Gibert, Nail, Arefyev, Ba{\~n}{\'o}n, van~der Linde, Ji, Zaragoza-Bernabeu, Aulamo, Ram{\'\i}rez-S{\'a}nchez, Kutuzov, Pyysalo, Oepen, and Tiedemann}]{de-gibert-etal-2024-new-massive}
Ona de~Gibert, Graeme Nail, Nikolay Arefyev, Marta Ba{\~n}{\'o}n, Jelmer van~der Linde, Shaoxiong Ji, Jaume Zaragoza-Bernabeu, Mikko Aulamo, Gema Ram{\'\i}rez-S{\'a}nchez, Andrey Kutuzov, Sampo Pyysalo, Stephan Oepen, and J{\"o}rg Tiedemann. 2024.
\newblock \href {https://aclanthology.org/2024.lrec-main.100} {A new massive multilingual dataset for high-performance language technologies}.
\newblock In \emph{Proceedings of the 2024 Joint International Conference on Computational Linguistics, Language Resources and Evaluation (LREC-COLING 2024)}, pages 1116--1128, Torino, Italia. ELRA and ICCL.

\bibitem[{de~Vries et~al.(2023)de~Vries, Wieling, and Nissim}]{de2023dumb}
Wietse de~Vries, Martijn Wieling, and Malvina Nissim. 2023.
\newblock Dumb: A benchmark for smart evaluation of dutch models.
\newblock \emph{arXiv preprint arXiv:2305.13026}.

\bibitem[{Devlin(2018)}]{devlin2018bert}
Jacob Devlin. 2018.
\newblock Bert: Pre-training of deep bidirectional transformers for language understanding.
\newblock \emph{arXiv preprint arXiv:1810.04805}.

\bibitem[{Devlin et~al.(2019)Devlin, Chang, Lee, and Toutanova}]{devlin-etal-2019-bert}
Jacob Devlin, Ming-Wei Chang, Kenton Lee, and Kristina Toutanova. 2019.
\newblock \href {https://doi.org/10.18653/v1/N19-1423} {{BERT}: Pre-training of deep bidirectional transformers for language understanding}.
\newblock In \emph{Proceedings of the 2019 Conference of the North {A}merican Chapter of the Association for Computational Linguistics: Human Language Technologies, Volume 1 (Long and Short Papers)}, pages 4171--4186, Minneapolis, Minnesota. Association for Computational Linguistics.

\bibitem[{Držík and Forgac(2024)}]{drzik2024slovak}
Dávid Držík and František Forgac. 2024.
\newblock \href {https://doi.org/10.7717/peerj-cs.2465} {Slovak morphological tokenizer using the byte-pair encoding algorithm}.
\newblock \emph{PeerJ Computer Science}, 10:e2465.

\bibitem[{Dumitrescu et~al.(2021)Dumitrescu, Rebeja, Lorincz, Gaman, Avram, Ilie, Pruteanu, Stan, Rosia, Iacobescu et~al.}]{dumitrescu2021liro}
Stefan~Daniel Dumitrescu, Petru Rebeja, Beata Lorincz, Mihaela Gaman, Andrei Avram, Mihai Ilie, Andrei Pruteanu, Adriana Stan, Lorena Rosia, Cristina Iacobescu, et~al. 2021.
\newblock Liro: Benchmark and leaderboard for romanian language tasks.
\newblock In \emph{Thirty-fifth Conference on Neural Information Processing Systems Datasets and Benchmarks Track (Round 1)}.

\bibitem[{Elmadany et~al.(2022)Elmadany, Nagoudi, and Abdul-Mageed}]{elmadany2022orca}
AbdelRahim Elmadany, El~Moatez~Billah Nagoudi, and Muhammad Abdul-Mageed. 2022.
\newblock Orca: A challenging benchmark for arabic language understanding.
\newblock \emph{arXiv preprint arXiv:2212.10758}.

\bibitem[{Ethayarajh and Jurafsky(2020)}]{ethayarajh-jurafsky-2020-utility}
Kawin Ethayarajh and Dan Jurafsky. 2020.
\newblock \href {https://doi.org/10.18653/v1/2020.emnlp-main.393} {Utility is in the eye of the user: A critique of {NLP} leaderboards}.
\newblock In \emph{Proceedings of the 2020 Conference on Empirical Methods in Natural Language Processing (EMNLP)}, pages 4846--4853, Online. Association for Computational Linguistics.

\bibitem[{Gajdo{\v s}ov{\'a} et~al.(2016)Gajdo{\v s}ov{\'a}, {\v S}imkov{\'a}, and et~al.}]{11234/1-1822}
Katar{\'{\i}}na Gajdo{\v s}ov{\'a}, M{\'a}ria {\v S}imkov{\'a}, and et~al. 2016.
\newblock \href {http://hdl.handle.net/11234/1-1822} {Slovak dependency treebank}.
\newblock {LINDAT}/{CLARIAH}-{CZ} digital library at the Institute of Formal and Applied Linguistics ({{\'U}FAL}), Faculty of Mathematics and Physics, Charles University.

\bibitem[{Giampiccolo et~al.(2007)Giampiccolo, Magnini, Dagan, and Dolan}]{giampiccolo2007third}
Danilo Giampiccolo, Bernardo Magnini, Ido Dagan, and William~B Dolan. 2007.
\newblock The third pascal recognizing textual entailment challenge.
\newblock In \emph{Proceedings of the ACL-PASCAL workshop on textual entailment and paraphrasing}, pages 1--9.

\bibitem[{Gurgurov et~al.(2025)Gurgurov, Kumar, and Ostermann}]{gurgurov2025gremlinrepositorygreenbaseline}
Daniil Gurgurov, Rishu Kumar, and Simon Ostermann. 2025.
\newblock \href {https://arxiv.org/abs/2409.18193} {Gremlin: A repository of green baseline embeddings for 87 low-resource languages injected with multilingual graph knowledge}.
\newblock \emph{Preprint}, arXiv:2409.18193.

\bibitem[{Hardalov et~al.(2023)Hardalov, Atanasova, Mihaylov, Angelova, Simov, Osenova, Stoyanov, Koychev, Nakov, and Radev}]{hardalov-etal-2023-bgglue}
Momchil Hardalov, Pepa Atanasova, Todor Mihaylov, Galia Angelova, Kiril Simov, Petya Osenova, Veselin Stoyanov, Ivan Koychev, Preslav Nakov, and Dragomir Radev. 2023.
\newblock \href {https://doi.org/10.18653/v1/2023.acl-long.487} {bg{GLUE}: A {B}ulgarian general language understanding evaluation benchmark}.
\newblock In \emph{Proceedings of the 61st Annual Meeting of the Association for Computational Linguistics (Volume 1: Long Papers)}, pages 8733--8759, Toronto, Canada. Association for Computational Linguistics.

\bibitem[{He et~al.(2021)He, Gao, and Chen}]{he2021debertav3}
Pengcheng He, Jianfeng Gao, and Weizhu Chen. 2021.
\newblock \href {https://arxiv.org/abs/2111.09543} {Debertav3: Improving deberta using electra-style pre-training with gradient-disentangled embedding sharing}.
\newblock \emph{Preprint}, arXiv:2111.09543.

\bibitem[{Hládek et~al.(2023)Hládek, Staš, Juhár, and Koctúr}]{10082887}
Daniel Hládek, Ján Staš, Jozef Juhár, and Tomáš Koctúr. 2023.
\newblock \href {https://doi.org/10.1109/ACCESS.2023.3262308} {Slovak dataset for multilingual question answering}.
\newblock \emph{IEEE Access}, 11:32869--32881.

\bibitem[{Hu et~al.(2020)Hu, Ruder, Siddhant, Neubig, Firat, and Johnson}]{hu2020xtreme}
Junjie Hu, Sebastian Ruder, Aditya Siddhant, Graham Neubig, Orhan Firat, and Melvin Johnson. 2020.
\newblock Xtreme: A massively multilingual multi-task benchmark for evaluating cross-lingual generalisation.
\newblock In \emph{International Conference on Machine Learning}, pages 4411--4421. PMLR.

\bibitem[{Joshi et~al.(2020)Joshi, Santy, Budhiraja, Bali, and Choudhury}]{joshi-etal-2020-state}
Pratik Joshi, Sebastin Santy, Amar Budhiraja, Kalika Bali, and Monojit Choudhury. 2020.
\newblock \href {https://doi.org/10.18653/v1/2020.acl-main.560} {The state and fate of linguistic diversity and inclusion in the {NLP} world}.
\newblock In \emph{Proceedings of the 58th Annual Meeting of the Association for Computational Linguistics}, pages 6282--6293, Online. Association for Computational Linguistics.

\bibitem[{Khashabi et~al.(2021)Khashabi, Cohan, Shakeri, Hosseini, Pezeshkpour, Alikhani, Aminnaseri, Bitaab, Brahman, Ghazarian et~al.}]{khashabi2021parsinlu}
Daniel Khashabi, Arman Cohan, Siamak Shakeri, Pedram Hosseini, Pouya Pezeshkpour, Malihe Alikhani, Moin Aminnaseri, Marzieh Bitaab, Faeze Brahman, Sarik Ghazarian, et~al. 2021.
\newblock Parsinlu: a suite of language understanding challenges for persian.
\newblock \emph{Transactions of the Association for Computational Linguistics}, 9:1147--1162.

\bibitem[{Kiela et~al.(2021)Kiela, Bartolo, Nie, Kaushik, Geiger, Wu, Vidgen, Prasad, Singh, Ringshia, Ma, Thrush, Riedel, Waseem, Stenetorp, Jia, Bansal, Potts, and Williams}]{kiela-etal-2021-dynabench}
Douwe Kiela, Max Bartolo, Yixin Nie, Divyansh Kaushik, Atticus Geiger, Zhengxuan Wu, Bertie Vidgen, Grusha Prasad, Amanpreet Singh, Pratik Ringshia, Zhiyi Ma, Tristan Thrush, Sebastian Riedel, Zeerak Waseem, Pontus Stenetorp, Robin Jia, Mohit Bansal, Christopher Potts, and Adina Williams. 2021.
\newblock \href {https://doi.org/10.18653/v1/2021.naacl-main.324} {Dynabench: Rethinking benchmarking in {NLP}}.
\newblock In \emph{Proceedings of the 2021 Conference of the North American Chapter of the Association for Computational Linguistics: Human Language Technologies}, pages 4110--4124, Online. Association for Computational Linguistics.

\bibitem[{Koppel and Ordan(2011)}]{koppel-ordan-2011-translationese}
Moshe Koppel and Noam Ordan. 2011.
\newblock \href {https://aclanthology.org/P11-1132/} {Translationese and its dialects}.
\newblock In \emph{Proceedings of the 49th Annual Meeting of the Association for Computational Linguistics: Human Language Technologies}, pages 1318--1326, Portland, Oregon, USA. Association for Computational Linguistics.

\bibitem[{Kudugunta et~al.(2024)Kudugunta, Caswell, Zhang, Garcia, Xin, Kusupati, Stella, Bapna, and Firat}]{kudugunta2024madlad}
Sneha Kudugunta, Isaac Caswell, Biao Zhang, Xavier Garcia, Derrick Xin, Aditya Kusupati, Romi Stella, Ankur Bapna, and Orhan Firat. 2024.
\newblock Madlad-400: A multilingual and document-level large audited dataset.
\newblock \emph{Advances in Neural Information Processing Systems}, 36.

\bibitem[{Kurihara et~al.(2022)Kurihara, Kawahara, and Shibata}]{kurihara2022jglue}
Kentaro Kurihara, Daisuke Kawahara, and Tomohide Shibata. 2022.
\newblock Jglue: Japanese general language understanding evaluation.
\newblock In \emph{Proceedings of the Thirteenth Language Resources and Evaluation Conference}, pages 2957--2966.

\bibitem[{Le et~al.(2019)Le, Vial, Frej, Segonne, Coavoux, Lecouteux, Allauzen, Crabb{\'e}, Besacier, and Schwab}]{le2019flaubert}
Hang Le, Lo{\"\i}c Vial, Jibril Frej, Vincent Segonne, Maximin Coavoux, Benjamin Lecouteux, Alexandre Allauzen, Benoit Crabb{\'e}, Laurent Besacier, and Didier Schwab. 2019.
\newblock Flaubert: Unsupervised language model pre-training for french.
\newblock \emph{arXiv preprint arXiv:1912.05372}.

\bibitem[{Lehe{\v{c}}ka and {\v{S}}vec(2021)}]{FERNETC5}
Jan Lehe{\v{c}}ka and Jan {\v{S}}vec. 2021.
\newblock \href {https://doi.org/10.1007/978-3-030-89579-2_3} {Comparison of czech transformers on text classification tasks}.
\newblock In \emph{Statistical Language and Speech Processing}, pages 27--37, Cham. Springer International Publishing.

\bibitem[{Lhoest et~al.(2021)Lhoest, Villanova~del Moral, Jernite, Thakur, von Platen, Patil, Chaumond, Drame, Plu, Tunstall, Davison, {\v{S}}a{\v{s}}ko, Chhablani, Malik, Brandeis, Le~Scao, Sanh, Xu, Patry, McMillan-Major, Schmid, Gugger, Delangue, Matussi{\`e}re, Debut, Bekman, Cistac, Goehringer, Mustar, Lagunas, Rush, and Wolf}]{lhoest-etal-2021-datasets}
Quentin Lhoest, Albert Villanova~del Moral, Yacine Jernite, Abhishek Thakur, Patrick von Platen, Suraj Patil, Julien Chaumond, Mariama Drame, Julien Plu, Lewis Tunstall, Joe Davison, Mario {\v{S}}a{\v{s}}ko, Gunjan Chhablani, Bhavitvya Malik, Simon Brandeis, Teven Le~Scao, Victor Sanh, Canwen Xu, Nicolas Patry, Angelina McMillan-Major, Philipp Schmid, Sylvain Gugger, Cl{\'e}ment Delangue, Th{\'e}o Matussi{\`e}re, Lysandre Debut, Stas Bekman, Pierric Cistac, Thibault Goehringer, Victor Mustar, Fran{\c{c}}ois Lagunas, Alexander Rush, and Thomas Wolf. 2021.
\newblock \href {https://doi.org/10.18653/v1/2021.emnlp-demo.21} {Datasets: A community library for natural language processing}.
\newblock In \emph{Proceedings of the 2021 Conference on Empirical Methods in Natural Language Processing: System Demonstrations}, pages 175--184, Online and Punta Cana, Dominican Republic. Association for Computational Linguistics.

\bibitem[{Liang et~al.(2023)Liang, Gonen, Mao, Hou, Goyal, Ghazvininejad, Zettlemoyer, and Khabsa}]{liang-etal-2023-xlm}
Davis Liang, Hila Gonen, Yuning Mao, Rui Hou, Naman Goyal, Marjan Ghazvininejad, Luke Zettlemoyer, and Madian Khabsa. 2023.
\newblock \href {https://doi.org/10.18653/v1/2023.emnlp-main.813} {{XLM}-{V}: Overcoming the vocabulary bottleneck in multilingual masked language models}.
\newblock In \emph{Proceedings of the 2023 Conference on Empirical Methods in Natural Language Processing}, pages 13142--13152, Singapore. Association for Computational Linguistics.

\bibitem[{Liang et~al.(2020)Liang, Duan, Gong, Wu, Guo, Qi, Gong, Shou, Jiang, Cao et~al.}]{liang2020xglue}
Yaobo Liang, Nan Duan, Yeyun Gong, Ning Wu, Fenfei Guo, Weizhen Qi, Ming Gong, Linjun Shou, Daxin Jiang, Guihong Cao, et~al. 2020.
\newblock Xglue: A new benchmark dataset for cross-lingual pre-training, understanding and generation.
\newblock \emph{arXiv preprint arXiv:2004.01401}.

\bibitem[{Liu et~al.(2019)Liu, Ott, Goyal, Du, Joshi, Chen, Levy, Lewis, Zettlemoyer, and Stoyanov}]{liu2019roberta}
Yinhan Liu, Myle Ott, Naman Goyal, Jingfei Du, Mandar Joshi, Danqi Chen, Omer Levy, Mike Lewis, Luke Zettlemoyer, and Veselin Stoyanov. 2019.
\newblock Roberta: A robustly optimized bert pretraining approach.
\newblock \emph{arXiv e-prints}, pages arXiv--1907.

\bibitem[{Loshchilov and Hutter(2017)}]{DBLP:journals/corr/abs-1711-05101}
Ilya Loshchilov and Frank Hutter. 2017.
\newblock \href {https://arxiv.org/abs/1711.05101} {Fixing weight decay regularization in adam}.
\newblock \emph{CoRR}, abs/1711.05101.

\bibitem[{Ma et~al.(2021)Ma, Ethayarajh, Thrush, Jain, Wu, Jia, Potts, Williams, and Kiela}]{ma2021dynaboard}
Zhiyi Ma, Kawin Ethayarajh, Tristan Thrush, Somya Jain, Ledell Wu, Robin Jia, Christopher Potts, Adina Williams, and Douwe Kiela. 2021.
\newblock Dynaboard: An evaluation-as-a-service platform for holistic next-generation benchmarking.
\newblock \emph{Advances in Neural Information Processing Systems}, 34:10351--10367.

\bibitem[{Mayhew et~al.(2024)Mayhew, Blevins, Liu, Suppa, Gonen, Imperial, Karlsson, Lin, Ljube{\v{s}}i{\'c}, Miranda, Plank, Riabi, and Pinter}]{mayhew-etal-2024-universal}
Stephen Mayhew, Terra Blevins, Shuheng Liu, Marek Suppa, Hila Gonen, Joseph~Marvin Imperial, B{\"o}rje Karlsson, Peiqin Lin, Nikola Ljube{\v{s}}i{\'c}, Lester~James Miranda, Barbara Plank, Arij Riabi, and Yuval Pinter. 2024.
\newblock \href {https://doi.org/10.18653/v1/2024.naacl-long.243} {Universal {NER}: A gold-standard multilingual named entity recognition benchmark}.
\newblock In \emph{Proceedings of the 2024 Conference of the North American Chapter of the Association for Computational Linguistics: Human Language Technologies (Volume 1: Long Papers)}, pages 4322--4337, Mexico City, Mexico. Association for Computational Linguistics.

\bibitem[{McHugh(2012)}]{mchugh2012interrater}
Mary~L McHugh. 2012.
\newblock Interrater reliability: the kappa statistic.
\newblock \emph{Biochemia medica}, 22(3):276--282.

\bibitem[{Nivre et~al.(2020)Nivre, de~Marneffe, Ginter, Haji{\v{c}}, Manning, Pyysalo, Schuster, Tyers, and Zeman}]{nivre-etal-2020-universal}
Joakim Nivre, Marie-Catherine de~Marneffe, Filip Ginter, Jan Haji{\v{c}}, Christopher~D. Manning, Sampo Pyysalo, Sebastian Schuster, Francis Tyers, and Daniel Zeman. 2020.
\newblock \href {https://aclanthology.org/2020.lrec-1.497/} {{U}niversal {D}ependencies v2: An evergrowing multilingual treebank collection}.
\newblock In \emph{Proceedings of the Twelfth Language Resources and Evaluation Conference}, pages 4034--4043, Marseille, France. European Language Resources Association.

\bibitem[{Os{\'o}rio et~al.(2024)Os{\'o}rio, Leite, Cardoso, Gomes, Rodrigues, Santos, and Branco}]{osorio2024portulan}
Tom{\'a}s Os{\'o}rio, Bernardo Leite, Henrique~Lopes Cardoso, Lu{\'\i}s Gomes, Jo{\~a}o Rodrigues, Rodrigo Santos, and Ant{\'o}nio Branco. 2024.
\newblock Portulan extraglue datasets and models: Kick-starting a benchmark for the neural processing of portuguese.
\newblock \emph{arXiv preprint arXiv:2404.05333}.

\bibitem[{Park(2021)}]{park2021klue}
Sungjoon Park. 2021.
\newblock Klue: Korean language understanding evaluation.
\newblock \emph{arXiv preprint arXiv:2105.09680}.

\bibitem[{Pecar et~al.(2019)Pecar, Simko, and Bielikova}]{pecar-etal-2019-improving}
Samuel Pecar, Marian Simko, and Maria Bielikova. 2019.
\newblock \href {https://doi.org/10.18653/v1/W19-3716} {Improving sentiment classification in {S}lovak language}.
\newblock In \emph{Proceedings of the 7th Workshop on Balto-Slavic Natural Language Processing}, pages 114--119, Florence, Italy. Association for Computational Linguistics.

\bibitem[{Pfister and Hotho(2024)}]{pfister2024supergleber}
Jan Pfister and Andreas Hotho. 2024.
\newblock Supergleber: German language understanding evaluation benchmark.
\newblock In \emph{Proceedings of the 2024 Conference of the North American Chapter of the Association for Computational Linguistics: Human Language Technologies (Volume 1: Long Papers)}, pages 7897--7916.

\bibitem[{Pikuliak et~al.(2021)Pikuliak, Grivalsk{\`y}, Kon{\^o}pka, Bl{\v{s}}t{\'a}k, Tamajka, Bachrat{\`y}, {\v{S}}imko, Bal{\'a}{\v{z}}ik, Trnka, and Uhl{\'a}rik}]{pikuliak2021slovakbert}
Mat{\'u}{\v{s}} Pikuliak, {\v{S}}tefan Grivalsk{\`y}, Martin Kon{\^o}pka, Miroslav Bl{\v{s}}t{\'a}k, Martin Tamajka, Viktor Bachrat{\`y}, Mari{\'a}n {\v{S}}imko, Pavol Bal{\'a}{\v{z}}ik, Michal Trnka, and Filip Uhl{\'a}rik. 2021.
\newblock Slovakbert: Slovak masked language model.
\newblock \emph{arXiv preprint arXiv:2109.15254}.

\bibitem[{Piskorski et~al.(2017)Piskorski, Pivovarova, {\v{S}}najder, Steinberger, and Yangarber}]{piskorski2017first}
Jakub Piskorski, Lidia Pivovarova, Jan {\v{S}}najder, Josef Steinberger, and Roman Yangarber. 2017.
\newblock The first cross-lingual challenge on recognition, normalization and matching of named entities in slavic languages.
\newblock In \emph{Workshop on Balto-Slavic Natural Language Processing}, pages 76--85. The Association for Computational Linguistics.

\bibitem[{Puccetti et~al.(2025)Puccetti, Cassese, and Esuli}]{puccetti-etal-2025-invalsi}
Giovanni Puccetti, Maria Cassese, and Andrea Esuli. 2025.
\newblock \href {https://aclanthology.org/2025.coling-main.453/} {The invalsi benchmarks: measuring the linguistic and mathematical understanding of large language models in {I}talian}.
\newblock In \emph{Proceedings of the 31st International Conference on Computational Linguistics}, pages 6782--6797, Abu Dhabi, UAE. Association for Computational Linguistics.

\bibitem[{Rodriguez-Penagos et~al.(2021)Rodriguez-Penagos, Armentano-Oller, Villegas, Melero, Gonzalez, Bonet, and Pio}]{rodriguez2021catalan}
Carlos Rodriguez-Penagos, Carme Armentano-Oller, Marta Villegas, Maite Melero, Aitor Gonzalez, Ona de~Gibert Bonet, and Casimiro~Carrino Pio. 2021.
\newblock The catalan language club.
\newblock \emph{arXiv preprint arXiv:2112.01894}.

\bibitem[{Ruder et~al.(2021)Ruder, Constant, Botha, Siddhant, Firat, Fu, Liu, Hu, Garrette, Neubig et~al.}]{ruder2021xtreme}
Sebastian Ruder, Noah Constant, Jan Botha, Aditya Siddhant, Orhan Firat, Jinlan Fu, Pengfei Liu, Junjie Hu, Dan Garrette, Graham Neubig, et~al. 2021.
\newblock Xtreme-r: Towards more challenging and nuanced multilingual evaluation.
\newblock \emph{arXiv preprint arXiv:2104.07412}.

\bibitem[{Rybak et~al.(2020)Rybak, Mroczkowski, Tracz, and Gawlik}]{rybak2020klej}
Piotr Rybak, Robert Mroczkowski, Janusz Tracz, and Ireneusz Gawlik. 2020.
\newblock Klej: Comprehensive benchmark for polish language understanding.
\newblock \emph{arXiv preprint arXiv:2005.00630}.

\bibitem[{Samuel et~al.(2023{\natexlab{a}})Samuel, Kutuzov, {\O}vrelid, and Velldal}]{samuel-etal-2023-trained}
David Samuel, Andrey Kutuzov, Lilja {\O}vrelid, and Erik Velldal. 2023{\natexlab{a}}.
\newblock \href {https://doi.org/10.18653/v1/2023.findings-eacl.146} {Trained on 100 million words and still in shape: {BERT} meets {B}ritish {N}ational {C}orpus}.
\newblock In \emph{Findings of the Association for Computational Linguistics: EACL 2023}, pages 1954--1974, Dubrovnik, Croatia. Association for Computational Linguistics.

\bibitem[{Samuel et~al.(2023{\natexlab{b}})Samuel, Kutuzov, Touileb, Velldal, {\O}vrelid, R{\o}nningstad, Sigdel, and Palatkina}]{samuel2023norbench}
David Samuel, Andrey Kutuzov, Samia Touileb, Erik Velldal, Lilja {\O}vrelid, Egil R{\o}nningstad, Elina Sigdel, and Anna Palatkina. 2023{\natexlab{b}}.
\newblock Norbench--a benchmark for norwegian language models.
\newblock \emph{arXiv preprint arXiv:2305.03880}.

\bibitem[{Sanh et~al.(2019)Sanh, Debut, Chaumond, and Wolf}]{Sanh2019DistilBERTAD}
Victor Sanh, Lysandre Debut, Julien Chaumond, and Thomas Wolf. 2019.
\newblock Distilbert, a distilled version of bert: smaller, faster, cheaper and lighter.
\newblock \emph{ArXiv}, abs/1910.01108.

\bibitem[{Shavrina et~al.(2020)Shavrina, Fenogenova, Emelyanov, Shevelev, Artemova, Malykh, Mikhailov, Tikhonova, Chertok, and Evlampiev}]{shavrina2020russiansuperglue}
Tatiana Shavrina, Alena Fenogenova, Anton Emelyanov, Denis Shevelev, Ekaterina Artemova, Valentin Malykh, Vladislav Mikhailov, Maria Tikhonova, Andrey Chertok, and Andrey Evlampiev. 2020.
\newblock Russiansuperglue: A russian language understanding evaluation benchmark.
\newblock \emph{arXiv preprint arXiv:2010.15925}.

\bibitem[{Strubell et~al.(2020)Strubell, Ganesh, and McCallum}]{strubell2020energy}
Emma Strubell, Ananya Ganesh, and Andrew McCallum. 2020.
\newblock Energy and policy considerations for modern deep learning research.
\newblock In \emph{Proceedings of the AAAI conference on artificial intelligence}, volume~34, pages 13693--13696.

\bibitem[{Suba et~al.(2023)Suba, Suppa, Kubik, Hamerlik, and Takac}]{suba-etal-2023-wikigoldsk}
David Suba, Marek Suppa, Jozef Kubik, Endre Hamerlik, and Martin Takac. 2023.
\newblock \href {https://doi.org/10.18653/v1/2023.bsnlp-1.16} {{W}iki{G}old{SK}: Annotated dataset, baselines and few-shot learning experiments for {S}lovak named entity recognition}.
\newblock In \emph{Proceedings of the 9th Workshop on Slavic Natural Language Processing 2023 (SlavicNLP 2023)}, pages 138--145, Dubrovnik, Croatia. Association for Computational Linguistics.

\bibitem[{Thrush et~al.(2022)Thrush, Tirumala, Gupta, Bartolo, Rodriguez, Kane, Gaviria~Rojas, Mattson, Williams, and Kiela}]{thrush-etal-2022-dynatask}
Tristan Thrush, Kushal Tirumala, Anmol Gupta, Max Bartolo, Pedro Rodriguez, Tariq Kane, William Gaviria~Rojas, Peter Mattson, Adina Williams, and Douwe Kiela. 2022.
\newblock \href {https://doi.org/10.18653/v1/2022.acl-demo.17} {Dynatask: A framework for creating dynamic {AI} benchmark tasks}.
\newblock In \emph{Proceedings of the 60th Annual Meeting of the Association for Computational Linguistics: System Demonstrations}, pages 174--181, Dublin, Ireland. Association for Computational Linguistics.

\bibitem[{Toral(2019)}]{toral-2019-post}
Antonio Toral. 2019.
\newblock \href {https://aclanthology.org/W19-6627/} {Post-editese: an exacerbated translationese}.
\newblock In \emph{Proceedings of Machine Translation Summit XVII: Research Track}, pages 273--281, Dublin, Ireland. European Association for Machine Translation.

\bibitem[{Urbizu et~al.(2022)Urbizu, San~Vicente, Saralegi, Agerri, and Soroa}]{urbizu2022basqueglue}
Gorka Urbizu, I{\~n}aki San~Vicente, Xabier Saralegi, Rodrigo Agerri, and Aitor Soroa. 2022.
\newblock Basqueglue: A natural language understanding benchmark for basque.
\newblock In \emph{Proceedings of the Thirteenth Language Resources and Evaluation Conference}, pages 1603--1612.

\bibitem[{Wang(2018)}]{wang2018glue}
Alex Wang. 2018.
\newblock Glue: A multi-task benchmark and analysis platform for natural language understanding.
\newblock \emph{arXiv preprint arXiv:1804.07461}.

\bibitem[{Wang et~al.(2019)Wang, Pruksachatkun, Nangia, Singh, Michael, Hill, Levy, and Bowman}]{wang2019superglue}
Alex Wang, Yada Pruksachatkun, Nikita Nangia, Amanpreet Singh, Julian Michael, Felix Hill, Omer Levy, and Samuel Bowman. 2019.
\newblock Superglue: A stickier benchmark for general-purpose language understanding systems.
\newblock \emph{Advances in neural information processing systems}, 32.

\bibitem[{Warner et~al.(2024)Warner, Chaffin, Clavié, Weller, Hallström, Taghadouini, Gallagher, Biswas, Ladhak, Aarsen, Cooper, Adams, Howard, and Poli}]{warner2024smarterbetterfasterlonger}
Benjamin Warner, Antoine Chaffin, Benjamin Clavié, Orion Weller, Oskar Hallström, Said Taghadouini, Alexis Gallagher, Raja Biswas, Faisal Ladhak, Tom Aarsen, Nathan Cooper, Griffin Adams, Jeremy Howard, and Iacopo Poli. 2024.
\newblock \href {https://arxiv.org/abs/2412.13663} {Smarter, better, faster, longer: A modern bidirectional encoder for fast, memory efficient, and long context finetuning and inference}.
\newblock \emph{Preprint}, arXiv:2412.13663.

\bibitem[{Wilie et~al.(2020)Wilie, Vincentio, Winata, Cahyawijaya, Li, Lim, Soleman, Mahendra, Fung, Bahar et~al.}]{wilie2020indonlu}
Bryan Wilie, Karissa Vincentio, Genta~Indra Winata, Samuel Cahyawijaya, Xiaohong Li, Zhi~Yuan Lim, Sidik Soleman, Rahmad Mahendra, Pascale Fung, Syafri Bahar, et~al. 2020.
\newblock Indonlu: Benchmark and resources for evaluating indonesian natural language understanding.
\newblock \emph{arXiv preprint arXiv:2009.05387}.

\bibitem[{Williams et~al.(2017)Williams, Nangia, and Bowman}]{williams2017broad}
Adina Williams, Nikita Nangia, and Samuel~R Bowman. 2017.
\newblock A broad-coverage challenge corpus for sentence understanding through inference.
\newblock \emph{arXiv preprint arXiv:1704.05426}.

\bibitem[{Wolf et~al.(2020)Wolf, Debut, Sanh, Chaumond, Delangue, Moi, Cistac, Rault, Louf, Funtowicz, Davison, Shleifer, von Platen, Ma, Jernite, Plu, Xu, Le~Scao, Gugger, Drame, Lhoest, and Rush}]{wolf-etal-2020-transformers}
Thomas Wolf, Lysandre Debut, Victor Sanh, Julien Chaumond, Clement Delangue, Anthony Moi, Pierric Cistac, Tim Rault, Remi Louf, Morgan Funtowicz, Joe Davison, Sam Shleifer, Patrick von Platen, Clara Ma, Yacine Jernite, Julien Plu, Canwen Xu, Teven Le~Scao, Sylvain Gugger, Mariama Drame, Quentin Lhoest, and Alexander Rush. 2020.
\newblock \href {https://doi.org/10.18653/v1/2020.emnlp-demos.6} {Transformers: State-of-the-art natural language processing}.
\newblock In \emph{Proceedings of the 2020 Conference on Empirical Methods in Natural Language Processing: System Demonstrations}, pages 38--45, Online. Association for Computational Linguistics.

\bibitem[{Xu et~al.(2020)Xu, Hu, Zhang, Li, Cao, Li, Xu, Sun, Yu, Yu, Tian, Dong, Liu, Shi, Cui, Li, Zeng, Wang, Xie, Li, Patterson, Tian, Zhang, Zhou, Liu, Zhao, Zhao, Yue, Zhang, Yang, Richardson, and Lan}]{xu-etal-2020-clue}
Liang Xu, Hai Hu, Xuanwei Zhang, Lu~Li, Chenjie Cao, Yudong Li, Yechen Xu, Kai Sun, Dian Yu, Cong Yu, Yin Tian, Qianqian Dong, Weitang Liu, Bo~Shi, Yiming Cui, Junyi Li, Jun Zeng, Rongzhao Wang, Weijian Xie, Yanting Li, Yina Patterson, Zuoyu Tian, Yiwen Zhang, He~Zhou, Shaoweihua Liu, Zhe Zhao, Qipeng Zhao, Cong Yue, Xinrui Zhang, Zhengliang Yang, Kyle Richardson, and Zhenzhong Lan. 2020.
\newblock \href {https://doi.org/10.18653/v1/2020.coling-main.419} {{CLUE}: A {C}hinese language understanding evaluation benchmark}.
\newblock In \emph{Proceedings of the 28th International Conference on Computational Linguistics}, pages 4762--4772, Barcelona, Spain (Online). International Committee on Computational Linguistics.

\bibitem[{{\v{Z}}agar and Robnik-{\v{S}}ikonja(2022)}]{vzagar2022slovene}
Ale{\v{s}} {\v{Z}}agar and Marko Robnik-{\v{S}}ikonja. 2022.
\newblock Slovene superglue benchmark: translation and evaluation.
\newblock \emph{arXiv preprint arXiv:2202.04994}.

\end{thebibliography}

\appendix

\newpage

\begin{table}[]
\centering
\begin{tabular}{l r}
\hline
\textbf{Model Name} & \textbf{\#Params} \\
\hline
XLM-R\textsubscript{Large} & 560M \\
XLM-R\textsubscript{Base}  & 278M \\
SlovakBERT                & 125M \\
mBERT\textsubscript{Base} & 178M \\
Distil-mBERT              & 135M \\
MiniLM\textsubscript{L12} & 118M \\
FERNET-CC\textsubscript{Base} & 164M\\
HPLT\textsubscript{Base} & 124M\\
mDeBERTaV3\textsubscript{Base} & 276M \\
DeBERTaV3\textsubscript{Base} & 184M \\
DeBERTaV3\textsubscript{Large} & 435M \\
ModernBERT\textsubscript{Base} & 149M \\
ModernBERT\textsubscript{Large} & 395M \\

\hline
\end{tabular}
\caption{The number of parameters in millions for each of the considered models.}
\label{tab:model_params}
\end{table}

\section{Training Setup and Hyperparameters}

\label{sec:hyper}

The main results presented here were obtained by evaluating 14 language models on 9 tasks (with the exception of the MobileBERT models on the QA task, which were unsupported at the time of writing). In total, we conducted 124 evaluation runs (including fine-tuning on the training set and inference on the test set). To determine the optimal hyperparameters, we performed an extensive search—detailed in the following pages—that encompassed 4,024 evaluation runs overall.

All models were trained on a single A100 (40GB) GPU using the following hyperparameters:

\begin{itemize} \item \textbf{Batch Size:} 12 \item \textbf{Weight Decay:} 0 \item \textbf{Learning Rate Decay:} Linear \item \textbf{Optimizer:} AdamW \cite{DBLP:journals/corr/abs-1711-05101} \item \textbf{Adam $\beta_1$:} 0.9 \item \textbf{Adam $\beta_2$:} 0.999 \item \textbf{Adam $\epsilon$:} 1e-8 \item \textbf{Gradient Clipping:} 1.0 \item \textbf{Dropout:} 0 \item \textbf{Epochs:} \textit{(see tables below)} \item \textbf{Warmup:} \textit{(see tables below)} \item \textbf{Learning Rate:} \textit{(see tables below)} \end{itemize}

Sweep durations (i.e., the total time per sweep, varying by task) are as follows:

\begin{itemize} \item \textbf{UD:} 100–720 minutes \item \textbf{UNER:} 130–500 minutes \item \textbf{WGSK:} 90–325 minutes \item \textbf{RTE:} 280–1850 minutes \item \textbf{NLI:} 2050–18300 minutes \item \textbf{STS:} 120–950 minutes \item \textbf{HS:} 560–5100 minutes \item \textbf{SA:} 100–630 minutes \item \textbf{QA:} 830–8400 minutes \end{itemize}

The average time per model run (computed as total time divided by the number of runs and models) is:

\begin{itemize} \item \textbf{UD:} 8 minutes \item \textbf{UNER:} 7 minutes \item \textbf{WGSK:} 5 minutes \item \textbf{RTE:} 12 minutes \item \textbf{NLI:} 4 hours 20 minutes \item \textbf{STS:} 10 minutes \item \textbf{HS:} 38 minutes \item \textbf{SA:} 7 minutes \item \textbf{QA:} 1 hour 15 minutes \end{itemize}

Overall, the experiments required on the order of 130 GPU days across all 9 tasks and 14 models.

\newpage

\enlargethispage{-\baselineskip}

\begin{table*}[h]

\centering
\resizebox{\textwidth}{!}{%
\begin{tabular}{p{0.2cm} l}
\toprule

\parbox[t]{0.2cm}{\rotatebox{90}{\textbf{UD}}} & \parbox{15cm}{
\textbf{Document:} \textit{Je potrebné chrániť bohatstvo lokálnych stredoamerických odrôd kukurice , pretože predstavujú zdroj biodiverzity pre ďalšie jej šľachtenie .}  / It is necessary to protect the wealth of local Central American corn varieties, because they represent a source of biodiversity for its further breeding. \\
\underline{\textbf{Tags:}} \underline{AUX}, \underline{ADJ}, \underline{VERB}, \underline{NOUN}, \underline{ADJ}, \underline{ADJ}, \underline{NOUN}, \underline{NOUN}, \underline{PUNCT}, \underline{SCONJ}, \underline{VERB}, \underline{NOUN}, \underline{NOUN}, \underline{ADP}, \underline{ADJ}, \underline{DET}, \underline{NOUN}, \underline{PUNCT} \\
} \\

\midrule

\parbox[t]{0.2cm}{\rotatebox{90}{\textbf{UNER}}} & \parbox{15cm}{
\textbf{Document:} \textit{V pakte medzi Hitlerom a Stalinom bolo Fínsko pridelené do sféry ZSSR .} / In the pact between Hitler and Stalin, Finland was assigned to the sphere of the USSR. \\
\underline{\textbf{Tags:}} \underline{O}, \underline{O}, \underline{O}, \underline{B-PER}, \underline{O}, \underline{B-PER}, \underline{O}, \underline{B-LOC}, \underline{O}, \underline{O}, \underline{O}, \underline{B-ORG}, \underline{O} \\
} \\

\midrule

\parbox[t]{0.2cm}{\rotatebox{90}{\textbf{WGSK}}} & \parbox{15cm}{
\textbf{Document:} \textit{Počas druhej svetovej vojny tu od roku 1939 do roku 1945 boli uskladnené umelecké zbierky Parížskeho múzea v Louvre .} / During the Second World War, from 1939 to 1945, the art collections of the Paris Louvre Museum were stored here. \\
\underline{\textbf{Tags:}} \underline{O}, \underline{B-MISC}, \underline{I-MISC}, \underline{I-MISC}, \underline{O}, \underline{O}, \underline{O}, \underline{O}, \underline{O}, \underline{O}, \underline{O}, \underline{O}, \underline{O}, \underline{O}, \underline{O}, \underline{B-ORG}, \underline{I-ORG}, \underline{I-ORG}, \underline{I-ORG}, \underline{O} \\
} \\

\midrule

\parbox[t]{0.2cm}{\rotatebox{90}{\textbf{RTE}}} & \parbox{15cm}{
\textbf{Text1:} \textit{Obrúsky, pozvánky a obyčajný starý papier stoja viac ako pred mesiacom.} / Tablecloths, invitations, and ordinary old paper cost more than they did a month ago. \\
\textbf{Text2:} \textit{Cena papiera rastie.} / The price of paper is rising. \\
\underline{\textbf{Correct Label:}} \underline{Entailment} \\
} \\

\midrule

\parbox[t]{0.2cm}{\rotatebox{90}{\textbf{NLI}}} & \parbox{15cm}{
\textbf{Premise:} \textit{Záblesky múdrosti by sa nemali prehliadať.} / Flashes of wisdom should not be overlooked. \\
\textbf{Hypothesis:} \textit{Záblesky múdrosti nie sú dôležité.} / Flashes of wisdom are not important. \\
\underline{\textbf{Entailment:}} \underline{Contradiction} \\
} \\

\bottomrule

\end{tabular}
}

\caption{English translations of the examples in Table \ref{tab:sklep-sample}. The original Slovak text are also included in the table for convenience and ease of comparison. For spacing reasons, the table continues on the next page.}
\label{tab:sklep-sample-translations}
\end{table*}

\newpage
\begin{table*}

\resizebox{\textwidth}{!}{%
\begin{tabular}{p{0.2cm} l}
\toprule

\parbox[t]{0.2cm}{\rotatebox{90}{\textbf{STS}}} & \parbox{15cm}{
\textbf{Premise:} \textit{Malý pes leží na posteli.} / A small dog is lying on the bed. \\
\textbf{Hypothesis:} \textit{Na posteli leží malý pes.} / A small dog lies on the bed. \\
\underline{\textbf{Similarity Score:}} \underline{5.0} \\
} \\

\midrule

\parbox[t]{0.2cm}{\rotatebox{90}{\textbf{HS}}} & \parbox{15cm}{
\textbf{Text:} \textit{Žiadna vláda kde budú feťáci a narkomani nebude nikdy dobrá} / No government where there are junkies and drug addicts will ever be good. \\
\underline{\textbf{Correct Label:}} \underline{Hate Speech} \\
} \\

\midrule

\parbox[t]{0.2cm}{\rotatebox{90}{\textbf{SA}}} & \parbox{15cm}{
\textbf{Text:} \textit{Pri vstupe do predajne Vás víta príjemný personál, čo mňa presvedčí o tom, že sem treba sa vracať aj druhýkrát, kde človek načerpá novú energiu do seba a samozrejme do svojho auta.} / At the entrance to the store, you are greeted by friendly staff, which convinces me that one should return here even a second time, where one can recharge oneself and, of course, one's car. \\
\underline{\textbf{Sentiment:}} \underline{Positive} \\
} \\

\midrule

\parbox[t]{0.2cm}{\rotatebox{90}{\textbf{QA}}} & \parbox{15cm}{
\textbf{Context:} \textit{Jozef Murgaš sa narodil v Tajove. Bol synom Jána Murgaša a Zuzany Murgašovej (rod. Slamovej). Základnú školu absolvoval v rodnom Tajove, neskôr študoval na gymnáziu v Banskej Bystrici (1876 – 1880), ale zaujímalo ho predovšetkým maliarstvo. V r. 1880 – 1882 študoval v bratislavskom seminári a neskôr do roku 1884 v ostrihomskom. ...} / Jozef Murgaš was born in Tajov. He was the son of Ján Murgaš and Zuzana Murgašová (née Slamová). He completed primary school in his hometown of Tajov, later he studied at the gymnasium in Banská Bystrica (1876 – 1880), but he was particularly interested in painting. From 1880 to 1882 he studied at the Bratislava seminary and later until 1884 at the Ostrihom seminary. ... \\
\textbf{Question:} \textit{Na akej škole študoval Jozef Murgaš v Banskej Bystrici ?} / At which school did Jozef Murgaš study in Banská Bystrica? \\
\underline{\textbf{Answer:}} \underline{na gymnáziu} / at the gymnasium
} \\

\bottomrule
    
\end{tabular}
}
\end{table*}

\newpage

\begin{table*}[]
\centering
\resizebox{\textwidth}{!}{%
\begin{tabular}{ll|lll|r}
\toprule
\textbf{Task} & \textbf{Model} & \textbf{Epochs} & \textbf{Warmup} & \textbf{LR} & \textbf{Dev} \\

\toprule

UD & XLM-R$_{Base}$ & \{1, 3, \textbf{5}\} & \{0.0, 0.1, \textbf{0.2}, 0.3\} & \{3e-05, \textbf{5e-05}, 0.0001\} & 97.67 \\
UD & XLM-R$_{Large}$ & \{1, 3, \textbf{5}\} & \{0.0, \textbf{0.1}, 0.2, 0.3\} & \{3e-05, \textbf{5e-05}, 0.0001\} & 97.99 \\
UD & HPLT$_{Base}$ & \{1, 3, \textbf{5}\} & \{0.0, 0.1, \textbf{0.2}, 0.3\} & \{3e-05, \textbf{5e-05}, 0.0001\} & 97.98 \\
UD & DistilmBERT$_{Base}$ & \{1, 3, \textbf{5}\} & \{0.0, \textbf{0.1}, 0.2, 0.3\} & \{3e-05, \textbf{5e-05}, 0.0001\} & 96.54 \\
UD & XLM-V$_{Base}$ & \{1, 3, \textbf{5}\} & \{0.0, 0.1, 0.2, \textbf{0.3}\} & \{3e-05, 5e-05, \textbf{0.0001}\} & 97.56 \\
UD & SlovakBERT & \{1, 3, \textbf{5}\} & \{\textbf{0.0}, 0.1, 0.2, 0.3\} & \{3e-05, \textbf{5e-05}, 0.0001\} & 97.86 \\
UD & mBERT$_{Base}$ & \{1, 3, \textbf{5}\} & \{\textbf{0.0}, 0.1, 0.2, 0.3\} & \{3e-05, \textbf{5e-05}, 0.0001\} & 96.26 \\
UD & DeBERTaV3$_{Large}$ & \{1, 3, \textbf{5}\} & \{\textbf{0.0}, 0.1, 0.2, 0.3\} & \{3e-05, \textbf{5e-05}, 0.0001\} & 97.21 \\
UD & mDeBERTaV3$_{Base}$ & \{1, 3, \textbf{5}\} & \{0.0, 0.1, \textbf{0.2}, 0.3\} & \{3e-05, 5e-05, \textbf{0.0001}\} & 97.68 \\
UD & FERNET-CC$_{Base}$ & \{1, 3, \textbf{5}\} & \{0.0, 0.1, \textbf{0.2}, 0.3\} & \{3e-05, \textbf{5e-05}, 0.0001\} & 97.92 \\
UD & DeBERTaV3$_{Base}$ & \{1, 3, \textbf{5}\} & \{0.0, \textbf{0.1}, 0.2, 0.3\} & \{3e-05, 5e-05, \textbf{0.0001}\} & 96.29 \\
UD & ModernBERT$_{Base}$ & \{1, 3, \textbf{5}\} & \{0.0, \textbf{0.1}, 0.2, 0.3\} & \{3e-05, 5e-05, \textbf{0.0001}\} & 89.80 \\
UD & ModernBERT$_{Large}$ & \{1, 3, \textbf{5}\} & \{0.0, \textbf{0.1}, 0.2, 0.3\} & \{3e-05, 5e-05, \textbf{0.0001}\} & 93.38 \\
UD & MiniLM$_{L12-Base}$ & \{1, 3, \textbf{5}\} & \{0.0, 0.1, 0.2, \textbf{0.3}\} & \{3e-05, 5e-05, \textbf{0.0001}\} & 96.92 \\

\midrule

UNER & XLM-R$_{Base}$ & \{1, 3, \textbf{5}\} & \{\textbf{0.0}, 0.1, 0.2, 0.3\} & \{1e-05, \textbf{3e-05}, 5e-05\} & 80.48 \\
UNER & XLM-R$_{Large}$ & \{1, 3, \textbf{5}\} & \{\textbf{0.0}, 0.1, 0.2, 0.3\} & \{\textbf{1e-05}, 3e-05, 5e-05\} & 85.12 \\
UNER & HPLT$_{Base}$ & \{1, 3, \textbf{5}\} & \{\textbf{0.0}, 0.1, 0.2, 0.3\} & \{1e-05, \textbf{3e-05}, 5e-05\} & 85.03 \\
UNER & DistilmBERT$_{Base}$ & \{1, \textbf{3}, 5\} & \{0.0, \textbf{0.1}, 0.2, 0.3\} & \{1e-05, \textbf{3e-05}, 5e-05\} & 77.75 \\
UNER & XLM-V$_{Base}$ & \{1, 3, \textbf{5}\} & \{0.0, \textbf{0.1}, 0.2, 0.3\} & \{1e-05, 3e-05, \textbf{5e-05}\} & 79.64 \\
UNER & SlovakBERT & \{1, \textbf{3}, 5\} & \{0.0, 0.1, 0.2, \textbf{0.3}\} & \{1e-05, \textbf{3e-05}, 5e-05\} & 82.57 \\
UNER & mBERT$_{Base}$ & \{1, 3, \textbf{5}\} & \{\textbf{0.0}, 0.1, 0.2, 0.3\} & \{1e-05, \textbf{3e-05}, 5e-05\} & 76.54 \\
UNER & DeBERTaV3$_{Large}$ & \{1, 3, \textbf{5}\} & \{0.0, 0.1, 0.2, \textbf{0.3}\} & \{\textbf{1e-05}, 3e-05, 5e-05\} & 73.34 \\
UNER & mDeBERTaV3$_{Base}$ & \{1, 3, \textbf{5}\} & \{0.0, 0.1, 0.2, \textbf{0.3}\} & \{1e-05, \textbf{3e-05}, 5e-05\} & 82.34 \\
UNER & FERNET-CC$_{Base}$ & \{1, \textbf{3}, 5\} & \{0.0, \textbf{0.1}, 0.2, 0.3\} & \{1e-05, 3e-05, \textbf{5e-05}\} & 83.67 \\
UNER & DeBERTaV3$_{Base}$ & \{1, 3, \textbf{5}\} & \{0.0, 0.1, \textbf{0.2}, 0.3\} & \{1e-05, 3e-05, \textbf{5e-05}\} & 72.81 \\
UNER & ModernBERT$_{Base}$ & \{1, 3, \textbf{5}\} & \{0.0, 0.1, \textbf{0.2}, 0.3\} & \{1e-05, 3e-05, \textbf{5e-05}\} & 47.19 \\
UNER & ModernBERT$_{Large}$ & \{1, 3, \textbf{5}\} & \{\textbf{0.0}, 0.1, 0.2, 0.3\} & \{1e-05, \textbf{3e-05}, 5e-05\} & 55.26 \\
UNER & MiniLM$_{L12-Base}$ & \{1, 3, \textbf{5}\} & \{0.0, 0.1, 0.2, \textbf{0.3}\} & \{1e-05, 3e-05, \textbf{5e-05}\} & 73.88 \\

\midrule

WGSK & XLM-R$_{Base}$ & \{1, 3, \textbf{5}\} & \{0.0, \textbf{0.1}, 0.2, 0.3\} & \{1e-05, 3e-05, \textbf{5e-05}\} & 91.37 \\
WGSK & XLM-R$_{Large}$ & \{1, 3, \textbf{5}\} & \{\textbf{0.0}, 0.1, 0.2, 0.3\} & \{1e-05, \textbf{3e-05}, 5e-05\} & 94.14 \\
WGSK & HPLT$_{Base}$ & \{1, 3, \textbf{5}\} & \{\textbf{0.0}, 0.1, 0.2, 0.3\} & \{1e-05, 3e-05, \textbf{5e-05}\} & 92.30 \\
WGSK & DistilmBERT$_{Base}$ & \{1, 3, \textbf{5}\} & \{0.0, 0.1, 0.2, \textbf{0.3}\} & \{1e-05, \textbf{3e-05}, 5e-05\} & 87.88 \\
WGSK & XLM-V$_{Base}$ & \{1, 3, \textbf{5}\} & \{0.0, 0.1, 0.2, \textbf{0.3}\} & \{1e-05, 3e-05, \textbf{5e-05}\} & 90.77 \\
WGSK & SlovakBERT & \{1, 3, \textbf{5}\} & \{0.0, \textbf{0.1}, 0.2, 0.3\} & \{1e-05, 3e-05, \textbf{5e-05}\} & 91.73 \\
WGSK & mBERT$_{Base}$ & \{1, 3, \textbf{5}\} & \{\textbf{0.0}, 0.1, 0.2, 0.3\} & \{1e-05, \textbf{3e-05}, 5e-05\} & 89.07 \\
WGSK & DeBERTaV3$_{Large}$ & \{1, 3, \textbf{5}\} & \{0.0, 0.1, 0.2, \textbf{0.3}\} & \{1e-05, \textbf{3e-05}, 5e-05\} & 90.91 \\
WGSK & mDeBERTaV3$_{Base}$ & \{1, 3, \textbf{5}\} & \{0.0, 0.1, \textbf{0.2}, 0.3\} & \{1e-05, 3e-05, \textbf{5e-05}\} & 92.80 \\
WGSK & FERNET-CC$_{Base}$ & \{1, 3, \textbf{5}\} & \{0.0, 0.1, 0.2, \textbf{0.3}\} & \{1e-05, 3e-05, \textbf{5e-05}\} & 92.57 \\
WGSK & DeBERTaV3$_{Base}$ & \{1, 3, \textbf{5}\} & \{0.0, \textbf{0.1}, 0.2, 0.3\} & \{1e-05, 3e-05, \textbf{5e-05}\} & 86.55 \\
WGSK & ModernBERT$_{Base}$ & \{1, 3, \textbf{5}\} & \{0.0, 0.1, 0.2, \textbf{0.3}\} & \{1e-05, 3e-05, \textbf{5e-05}\} & 66.07 \\
WGSK & ModernBERT$_{Large}$ & \{1, 3, \textbf{5}\} & \{0.0, 0.1, 0.2, \textbf{0.3}\} & \{1e-05, 3e-05, \textbf{5e-05}\} & 77.49 \\
WGSK & MiniLM$_{L12-Base}$ & \{1, 3, \textbf{5}\} & \{\textbf{0.0}, 0.1, 0.2, 0.3\} & \{1e-05, 3e-05, \textbf{5e-05}\} & 66.75 \\

\bottomrule
\end{tabular}}
\end{table*}

\begin{table*}[]
\centering
\resizebox{\textwidth}{!}{%
\begin{tabular}{ll|lll|r}
\toprule
\textbf{Task} & \textbf{Model} & \textbf{Epochs} & \textbf{Warmup} & \textbf{LR} & \textbf{Dev} \\

\toprule
\midrule

RTE & XLM-R$_{Base}$ & \{1, 3, \textbf{5}, 10\} & \{\textbf{0.0}, 0.1, 0.2, 0.3\} & \{1e-05, \textbf{3e-05}, 5e-05, 0.0001\} & 66.06 \\
RTE & XLM-R$_{Large}$ & \{1, 3, 5, \textbf{10}\} & \{0.0, 0.1, 0.2, \textbf{0.3}\} & \{1e-05, \textbf{3e-05}, 5e-05, 0.0001\} & 80.87 \\
RTE & HPLT$_{Base}$ & \{1, 3, \textbf{5}, 10\} & \{0.0, \textbf{0.1}, 0.2, 0.3\} & \{1e-05, 3e-05, \textbf{5e-05}, 0.0001\} & 57.40 \\
RTE & DistilmBERT$_{Base}$ & \{1, 3, 5, \textbf{10}\} & \{\textbf{0.0}, 0.1, 0.2, 0.3\} & \{1e-05, \textbf{3e-05}, 5e-05, 0.0001\} & 65.70 \\
RTE & XLM-V$_{Base}$ & \{1, 3, 5, \textbf{10}\} & \{0.0, \textbf{0.1}, 0.2, 0.3\} & \{\textbf{1e-05}, 3e-05, 5e-05, 0.0001\} & 65.34 \\
RTE & SlovakBERT & \{1, 3, \textbf{5}, 10\} & \{0.0, 0.1, \textbf{0.2}, 0.3\} & \{\textbf{1e-05}, 3e-05, 5e-05, 0.0001\} & 68.59 \\
RTE & mBERT$_{Base}$ & \{1, 3, \textbf{5}, 10\} & \{0.0, 0.1, \textbf{0.2}, 0.3\} & \{1e-05, \textbf{3e-05}, 5e-05, 0.0001\} & 71.84 \\
RTE & DeBERTaV3$_{Large}$ & \{1, 3, \textbf{5}, 10\} & \{0.0, \textbf{0.1}, 0.2, 0.3\} & \{1e-05, \textbf{3e-05}, 5e-05, 0.0001\} & 83.03 \\
RTE & mDeBERTaV3$_{Base}$ & \{1, 3, \textbf{5}, 10\} & \{0.0, \textbf{0.1}, 0.2, 0.3\} & \{1e-05, 3e-05, \textbf{5e-05}, 0.0001\} & 76.90 \\
RTE & FERNET-CC$_{Base}$ & \{1, 3, \textbf{5}, 10\} & \{0.0, \textbf{0.1}, 0.2, 0.3\} & \{1e-05, 3e-05, \textbf{5e-05}, 0.0001\} & 72.20 \\
RTE & DeBERTaV3$_{Base}$ & \{1, 3, \textbf{5}, 10\} & \{0.0, 0.1, \textbf{0.2}, 0.3\} & \{1e-05, \textbf{3e-05}, 5e-05, 0.0001\} & 67.15 \\
RTE & ModernBERT$_{Base}$ & \{1, 3, 5, \textbf{10}\} & \{0.0, 0.1, \textbf{0.2}, 0.3\} & \{1e-05, 3e-05, \textbf{5e-05}, 0.0001\} & 61.01 \\
RTE & ModernBERT$_{Large}$ & \{1, \textbf{3}, 5, 10\} & \{0.0, \textbf{0.1}, 0.2, 0.3\} & \{1e-05, 3e-05, 5e-05, \textbf{0.0001}\} & 64.26 \\
RTE & MiniLM$_{L12-Base}$ & \{1, 3, 5, \textbf{10}\} & \{\textbf{0.0}, 0.1, 0.2, 0.3\} & \{1e-05, 3e-05, \textbf{5e-05}, 0.0001\} & 71.12 \\

\midrule

NLI & XLM-R$_{Base}$ & \{1, \textbf{3}\} & \{\textbf{0.0}, 0.1, 0.2, 0.3\} & \{\textbf{1e-05}, 3e-05, 5e-05\} & 81.73 \\
NLI & DistilmBERT$_{Base}$ & \{1, \textbf{3}\} & \{0.0, 0.1, 0.2, \textbf{0.3}\} & \{1e-05, 3e-05, \textbf{5e-05}\} & 73.94 \\
NLI & FERNET-CC$_{Base}$ & \{\textbf{1}, 3\} & \{\textbf{0.0}, 0.1, 0.2, 0.3\} & \{1e-05, \textbf{3e-05}, 5e-05\} & 81.37 \\
NLI & SlovakBERT & \{1, \textbf{3}\} & \{0.0, \textbf{0.1}, 0.2, 0.3\} & \{1e-05, \textbf{3e-05}, 5e-05\} & 83.49 \\
NLI & mBERT$_{Base}$ & \{1, \textbf{3}\} & \{0.0, \textbf{0.1}, 0.2, 0.3\} & \{\textbf{1e-05}, 3e-05, 5e-05\} & 78.67 \\
NLI & XLM-R$_{Large}$ & \{\textbf{1}\} & \{0.0, 0.1, \textbf{0.2}, 0.3\} & \{\textbf{1e-05}, 3e-05, 5e-05\} & 86.91 \\
NLI & HPLT$_{Base}$ & \{\textbf{1}\} & \{\textbf{0.0}, 0.1, 0.2, 0.3\} & \{1e-05, 3e-05, \textbf{5e-05}\} & 81.37 \\
NLI & XLM-V$_{Base}$ & \{\textbf{1}\} & \{0.0, 0.1, \textbf{0.2}, 0.3\} & \{1e-05, \textbf{3e-05}, 5e-05\} & 80.24 \\
NLI & DeBERTaV3$_{Large}$ & \{\textbf{1}\} & \{\textbf{0.0}, 0.1, 0.2, 0.3\} & \{\textbf{1e-05}, 3e-05, 5e-05\} & 85.66 \\
NLI & mDeBERTaV3$_{Base}$ & \{\textbf{1}\} & \{\textbf{0.0}, 0.1, 0.2, 0.3\} & \{1e-05, \textbf{3e-05}, 5e-05\} & 84.98 \\
NLI & DeBERTaV3$_{Base}$ & \{1, \textbf{3}\} & \{0.0, \textbf{0.1}, 0.2, 0.3\} & \{\textbf{1e-05}, 3e-05, 5e-05\} & 74.82 \\
NLI & ModernBERT$_{Base}$ & \{1, \textbf{3}\} & \{\textbf{0.0}, 0.1, 0.2, 0.3\} & \{\textbf{1e-05}, 3e-05, 5e-05\} & 66.95 \\
NLI & ModernBERT$_{Large}$ & \{\textbf{1}\} & \{0.0, 0.1, \textbf{0.2}, 0.3\} & \{\textbf{1e-05}, 3e-05, 5e-05\} & 70.88 \\
NLI & MiniLM$_{L12-Base}$ & \{\textbf{1}\} & \{0.0, \textbf{0.1}, 0.2, 0.3\} & \{\textbf{1e-05}, 3e-05, 5e-05\} & 71.20 \\

\midrule
STS & XLM-R$_{Base}$ & \{1, 3, \textbf{5}\} & \{\textbf{0.0}, 0.1, 0.2, 0.3\} & \{1e-05, \textbf{3e-05}, 5e-05\} & 85.49 \\
STS & XLM-R$_{Large}$ & \{1, 3, \textbf{5}\} & \{0.0, 0.1, \textbf{0.2}, 0.3\} & \{\textbf{1e-05}, 3e-05, 5e-05\} & 89.06 \\
STS & HPLT$_{Base}$ & \{1, 3, \textbf{5}\} & \{0.0, 0.1, 0.2, \textbf{0.3}\} & \{1e-05, 3e-05, \textbf{5e-05}\} & 85.11 \\
STS & DistilmBERT$_{Base}$ & \{1, 3, \textbf{5}\} & \{0.0, \textbf{0.1}, 0.2, 0.3\} & \{1e-05, 3e-05, \textbf{5e-05}\} & 79.64 \\
STS & XLM-V$_{Base}$ & \{1, 3, \textbf{5}\} & \{\textbf{0.0}, 0.1, 0.2, 0.3\} & \{1e-05, \textbf{3e-05}, 5e-05\} & 84.49 \\
STS & SlovakBERT & \{1, 3, \textbf{5}\} & \{\textbf{0.0}, 0.1, 0.2, 0.3\} & \{1e-05, \textbf{3e-05}, 5e-05\} & 86.30 \\
STS & mBERT$_{Base}$ & \{1, 3, \textbf{5}\} & \{0.0, \textbf{0.1}, 0.2, 0.3\} & \{1e-05, \textbf{3e-05}, 5e-05\} & 85.02 \\
STS & DeBERTaV3$_{Large}$ & \{1, 3, \textbf{5}\} & \{0.0, \textbf{0.1}, 0.2, 0.3\} & \{1e-05, \textbf{3e-05}, 5e-05\} & 86.34 \\
STS & mDeBERTaV3$_{Base}$ & \{1, 3, \textbf{5}\} & \{0.0, 0.1, \textbf{0.2}, 0.3\} & \{1e-05, 3e-05, \textbf{5e-05}\} & 87.55 \\
STS & FERNET-CC$_{Base}$ & \{1, 3, \textbf{5}\} & \{\textbf{0.0}, 0.1, 0.2, 0.3\} & \{1e-05, 3e-05, \textbf{5e-05}\} & 87.84 \\
STS & DeBERTaV3$_{Base}$ & \{1, 3, \textbf{5}\} & \{0.0, 0.1, \textbf{0.2}, 0.3\} & \{1e-05, \textbf{3e-05}, 5e-05\} & 80.54 \\
STS & ModernBERT$_{Base}$ & \{1, 3, \textbf{5}\} & \{0.0, \textbf{0.1}, 0.2, 0.3\} & \{1e-05, 3e-05, \textbf{5e-05}\} & 80.96 \\
STS & ModernBERT$_{Large}$ & \{1, \textbf{3}, 5\} & \{0.0, 0.1, \textbf{0.2}, 0.3\} & \{1e-05, 3e-05, \textbf{5e-05}\} & 82.63 \\
STS & MiniLM$_{L12-Base}$ & \{1, 3, \textbf{5}\} & \{0.0, 0.1, 0.2, \textbf{0.3}\} & \{1e-05, 3e-05, \textbf{5e-05}\} & 83.30 \\

\bottomrule
\end{tabular}}
\end{table*}

\begin{table*}[]
\centering
\resizebox{\textwidth}{!}{%
\begin{tabular}{ll|lll|r}
\toprule
\textbf{Task} & \textbf{Model} & \textbf{Epochs} & \textbf{Warmup} & \textbf{LR} & \textbf{Dev} \\

\toprule

HS & XLM-R$_{Base}$ & \{1, \textbf{3}, 5, 10\} & \{0.0, 0.1, 0.2, \textbf{0.3}\} & \{\textbf{3e-05}, 5e-05, 0.0001\} & 82.30 \\
HS & XLM-R$_{Large}$ & \{\textbf{1}, 3, 5, 10\} & \{0.0, 0.1, \textbf{0.2}, 0.3\} & \{\textbf{3e-05}, 5e-05, 0.0001\} & 82.90 \\
HS & HPLT$_{Base}$ & \{1, \textbf{3}, 5, 10\} & \{0.0, 0.1, \textbf{0.2}, 0.3\} & \{3e-05, \textbf{5e-05}, 0.0001\} & 83.87 \\
HS & DistilmBERT$_{Base}$ & \{1, \textbf{3}, 5, 10\} & \{0.0, 0.1, \textbf{0.2}, 0.3\} & \{\textbf{3e-05}, 5e-05, 0.0001\} & 78.72 \\
HS & XLM-V$_{Base}$ & \{1, 3, \textbf{5}, 10\} & \{0.0, 0.1, \textbf{0.2}, 0.3\} & \{\textbf{3e-05}, 5e-05, 0.0001\} & 80.66 \\
HS & SlovakBERT & \{\textbf{1}, 3, 5, 10\} & \{0.0, \textbf{0.1}, 0.2, 0.3\} & \{\textbf{3e-05}, 5e-05, 0.0001\} & 85.74 \\
HS & mBERT$_{Base}$ & \{1, \textbf{3}, 5, 10\} & \{0.0, 0.1, 0.2, \textbf{0.3}\} & \{\textbf{3e-05}, 5e-05, 0.0001\} & 80.66 \\
HS & DeBERTaV3$_{Large}$ & \{1, 3, \textbf{5}, 10\} & \{0.0, 0.1, 0.2, \textbf{0.3}\} & \{\textbf{3e-05}, 5e-05, 0.0001\} & 81.33 \\
HS & mDeBERTaV3$_{Base}$ & \{1, \textbf{3}, 5, 10\} & \{0.0, 0.1, 0.2, \textbf{0.3}\} & \{3e-05, \textbf{5e-05}, 0.0001\} & 83.79 \\
HS & FERNET-CC$_{Base}$ & \{\textbf{1}, 3, 5, 10\} & \{0.0, \textbf{0.1}, 0.2, 0.3\} & \{3e-05, \textbf{5e-05}, 0.0001\} & 82.75 \\
HS & DeBERTaV3$_{Base}$ & \{1, 3, \textbf{5}, 10\} & \{0.0, 0.1, 0.2, \textbf{0.3}\} & \{3e-05, 5e-05, \textbf{0.0001}\} & 80.43 \\
HS & ModernBERT$_{Base}$ & \{1, \textbf{3}, 5, 10\} & \{0.0, 0.1, \textbf{0.2}, 0.3\} & \{3e-05, \textbf{5e-05}, 0.0001\} & 77.07 \\
HS & ModernBERT$_{Large}$ & \{1, 3, 5, \textbf{10}\} & \{0.0, 0.1, 0.2, \textbf{0.3}\} & \{3e-05, \textbf{5e-05}, 0.0001\} & 77.22 \\
HS & MiniLM$_{L12-Base}$ & \{1, 3, \textbf{5}, 10\} & \{0.0, \textbf{0.1}, 0.2, 0.3\} & \{\textbf{3e-05}, 5e-05, 0.0001\} & 79.76 \\

\midrule

SA & XLM-R$_{Base}$ & \{1, 3, \textbf{5}\} & \{0.0, 0.1, 0.2, \textbf{0.3}\} & \{\textbf{3e-05}, 5e-05, 0.0001\} & 98.28 \\
SA & XLM-R$_{Large}$ & \{1, 3, \textbf{5}\} & \{\textbf{0.0}, 0.1, 0.2, 0.3\} & \{\textbf{3e-05}, 5e-05, 0.0001\} & 98.66 \\
SA & HPLT$_{Base}$ & \{1, \textbf{3}, 5\} & \{0.0, 0.1, 0.2, \textbf{0.3}\} & \{3e-05, 5e-05, \textbf{0.0001}\} & 98.28 \\
SA & DistilmBERT$_{Base}$ & \{1, 3, \textbf{5}\} & \{0.0, 0.1, 0.2, \textbf{0.3}\} & \{3e-05, \textbf{5e-05}, 0.0001\} & 95.98 \\
SA & XLM-V$_{Base}$ & \{1, 3, \textbf{5}\} & \{0.0, \textbf{0.1}, 0.2, 0.3\} & \{3e-05, \textbf{5e-05}, 0.0001\} & 98.08 \\
SA & SlovakBERT & \{1, \textbf{3}, 5\} & \{\textbf{0.0}, 0.1, 0.2, 0.3\} & \{3e-05, \textbf{5e-05}, 0.0001\} & 98.08 \\
SA & mBERT$_{Base}$ & \{1, 3, \textbf{5}\} & \{0.0, 0.1, \textbf{0.2}, 0.3\} & \{3e-05, 5e-05, \textbf{0.0001}\} & 97.70 \\
SA & DeBERTaV3$_{Large}$ & \{1, 3, \textbf{5}\} & \{0.0, \textbf{0.1}, 0.2, 0.3\} & \{\textbf{3e-05}, 5e-05, 0.0001\} & 98.28 \\
SA & mDeBERTaV3$_{Base}$ & \{1, \textbf{3}, 5\} & \{0.0, 0.1, 0.2, \textbf{0.3}\} & \{\textbf{3e-05}, 5e-05, 0.0001\} & 99.04 \\
SA & FERNET-CC$_{Base}$ & \{1, 3, \textbf{5}\} & \{\textbf{0.0}, 0.1, 0.2, 0.3\} & \{3e-05, \textbf{5e-05}, 0.0001\} & 98.85 \\
SA & DeBERTaV3$_{Base}$ & \{1, 3, \textbf{5}\} & \{\textbf{0.0}, 0.1, 0.2, 0.3\} & \{3e-05, 5e-05, \textbf{0.0001}\} & 95.79 \\
SA & ModernBERT$_{Base}$ & \{1, 3, \textbf{5}\} & \{0.0, 0.1, \textbf{0.2}, 0.3\} & \{3e-05, 5e-05, \textbf{0.0001}\} & 93.68 \\
SA & ModernBERT$_{Large}$ & \{\textbf{1}, 3, 5\} & \{0.0, 0.1, \textbf{0.2}, 0.3\} & \{3e-05, \textbf{5e-05}, 0.0001\} & 95.02 \\
SA & MiniLM$_{L12-Base}$ & \{1, 3, \textbf{5}\} & \{\textbf{0.0}, 0.1, 0.2, 0.3\} & \{3e-05, \textbf{5e-05}, 0.0001\} & 96.36 \\

\midrule

QA & XLM-R$_{Base}$ & \{1, 2, \textbf{3}\} & \{0.0, 0.1, 0.2, \textbf{0.3}\} & \{3e-05, \textbf{5e-05}, 0.0001\} & 76.13 \\
QA & XLM-R$_{Large}$ & \{1, 2, \textbf{3}\} & \{0.0, 0.1, 0.2, \textbf{0.3}\} & \{\textbf{3e-05}, 5e-05, 0.0001\} & 79.88 \\
QA & HPLT$_{Base}$ & \{1, 2, \textbf{3}\} & \{0.0, 0.1, 0.2, \textbf{0.3}\} & \{3e-05, 5e-05, \textbf{0.0001}\} & 78.44 \\
QA & DistilmBERT$_{Base}$ & \{1, 2, \textbf{3}\} & \{0.0, 0.1, 0.2, \textbf{0.3}\} & \{3e-05, 5e-05, \textbf{0.0001}\} & 70.68 \\
QA & XLM-V$_{Base}$ & \{1, 2, \textbf{3}\} & \{\textbf{0.0}, 0.1, 0.2, 0.3\} & \{3e-05, \textbf{5e-05}, 0.0001\} & 76.13 \\
QA & SlovakBERT & \{1, 2, \textbf{3}\} & \{0.0, 0.1, \textbf{0.2}, 0.3\} & \{3e-05, \textbf{5e-05}, 0.0001\} & 77.40 \\
QA & mBERT$_{Base}$ & \{1, 2, \textbf{3}\} & \{0.0, \textbf{0.1}, 0.2, 0.3\} & \{\textbf{3e-05}, 5e-05, 0.0001\} & 76.42 \\
QA & DeBERTaV3$_{Large}$ & \{1, 2, \textbf{3}\} & \{0.0, 0.1, \textbf{0.2}, 0.3\} & \{\textbf{3e-05}, 5e-05, 0.0001\} & 77.69 \\
QA & mDeBERTaV3$_{Base}$ & \{1, 2, \textbf{3}\} & \{\textbf{0.0}, 0.1, 0.2, 0.3\} & \{\textbf{3e-05}, 5e-05, 0.0001\} & 78.65 \\
QA & FERNET-CC$_{Base}$ & \{1, \textbf{2}, 3\} & \{\textbf{0.0}, 0.1, 0.2, 0.3\} & \{3e-05, \textbf{5e-05}, 0.0001\} & 77.56 \\
QA & DeBERTaV3$_{Base}$ & \{1, 2, \textbf{3}\} & \{0.0, 0.1, \textbf{0.2}, 0.3\} & \{3e-05, \textbf{5e-05}, 0.0001\} & 74.41 \\

QA & MiniLM$_{L12-Base}$ & \{1, 2, \textbf{3}\} & \{0.0, 0.1, 0.2, \textbf{0.3}\} & \{3e-05, 5e-05, \textbf{0.0001}\} & 75.24 \\
\bottomrule
\end{tabular}}
\end{table*}

\onecolumn

\section{Relabeling experiment}
\label{sec:relabel}
We randomly sampled 100 examples from the training sets of the automatically translated tasks. Slovak native speakers (not involved in the test set post-editing) provided new labels. The results for each task are detailed below.

\noindent\rule{\linewidth}{0.4pt}

\subsection*{RTE}

A single annotator re-labeled 100 samples, leading to 11 discrepancies relative to the original labels. This yields a Cohen’s kappa of 0.77, indicating substantial agreement (as per \citep{mchugh2012interrater}). A second annotator (involved in post-editing) categorized these discrepancies as follows:

\begin{table}[h]
\centering
\begin{tabular}{l r}
\toprule
Category & Count\\
\midrule
Annotator Error & 6\\
Wrong Label & 3\\
Translation Error & 2\\
\bottomrule
\end{tabular}
\end{table}

With the categories defined as follows:
\begin{itemize}
\item \textbf{Annotation Error:} The original annotator selected an incorrect label.
\item \textbf{Wrong Label:} The original (English) dataset likely contained an incorrect label.
\item \textbf{Translation Error:} The label changed due to translation quality issues.
\end{itemize}

\textbf{Translation Issues}

We present the encountered translation issues below:
\begin{table}[h]
\centering
\begin{tabular}{p{0.45\linewidth} p{0.45\linewidth}}
\toprule
Original & Translation\\
\midrule
A man suspected of stealing a million-dollar collection of Nepalese and Tibetan art objects in New York was arrested. &
V New Yorku zatkli muža podozrivého z krádeže zbierky nepálskych a tibetských umeleckých predmetov v hodnote milión dolárov. (\emph{In New York, a man suspected of stealing a collection of Nepali and Tibetan artistic objects worth a million dollars was arrested.})\\[0.5em]
Free Speech is a part of the CBS Evening News. &
Slobodný prejav je súčasťou večerných správ CBS. (\emph{Free expression is part of the CBS’s evening news.})\\
\bottomrule
\end{tabular}
\end{table}

\begin{itemize}
\item In the first example, the translation incorrectly implies the arrest occurred in New York, the location of the robbery.
\item In the second, “Free Speech” was back-translated as “Free expression,” a term with distinct meanings in Slovak, leading to a label change.
\end{itemize}

Overall, the results suggests that the translation-induced error on the training set is on the order of single-digit percent (in this case 2\%).

\noindent\rule{\linewidth}{0.4pt}

\subsection*{NLI}

We applied the same two-step approach for NLI. The initial annotation produced 26 differing labels, with a Cohen’s kappa of 0.61, indicating substantial agreement [0]. The subsequent categorization is as follows:

\begin{table}[H]
\centering
\begin{tabular}{l r}
\toprule
Category & Count\\
\midrule
Annotator Error & 15\\
Wrong Label & 6\\
Translation Error & 5\\
\bottomrule
\end{tabular}
\end{table}

\textbf{Notable Translation Issues}

A selection of notable translation issues is described below:
\begin{table}[H]
\centering
\begin{tabular}{p{0.45\linewidth} p{0.45\linewidth}}
\toprule
Original & Translation\\
\midrule
The Indian population did not decline because of suicide . &
Populácia Indie neklesla kvôli samovraždám . (\emph{India's population did not decline because of suicides.})\\[0.5em]
We no longer offer the Singapore specials . &
Už neponúkame špeciálne ponuky pre Singapur . (\emph{We no longer offer special offers for Singapore.})\\
\bottomrule
\end{tabular}
\end{table}

\begin{itemize}
\item In the first sentence, “Indian population” was rendered as “India’s population” rather than the intended “Native American population” or “population of Indigenous peoples.”
\item In the second, “Singapore specials” were back-translated as “special offers for Singapore,” which alters the intended meaning and the corresponding label.
\end{itemize}

Note that a higher number of “wrong labels” is expected, given the inherent noisiness of the original dataset (e.g., the text of one sample was simply "n / a"). Moreover, translation was performed using an open-weights model of lower quality (MADLAD-400-3B). Nonetheless, translation error remains in the single-digit range (approximately 5\%).

\noindent\rule{\linewidth}{0.4pt}

\subsection*{STS}

For the STS task, we attempted to replicate the original methodology: similarity scores from three annotators were averaged to produce a final value, which was then compared to the original score using Mean Absolute Error (MAE), yielding 0.69. Annotators identified four translation issues that significantly impacted downstream scores, while other error types were deemed negligible due to the task’s subjectivity. Notably, most translation errors appeared among the top five samples with the highest absolute error relative to the original score.

\textbf{Notable Translation Issues}

\begin{table}[h]
\centering
\begin{tabular}{p{0.45\linewidth} p{0.45\linewidth}}
\toprule
Original & Translation\\
\midrule
Two hockey players in a struggle on the ice. &
Dvaja hokejisti v zápase na ľade. (\emph{Two hockey players in a match on ice.})\\[0.5em]
Nelson Mandela dies: Live coverage &
Nelson Mandela zomiera: Priamy prenos. (\emph{Nelson Mandela is dying: Live broadcast})\\
\bottomrule
\end{tabular}
\end{table}

\begin{itemize}
\item In the first case, the move from "struggle" to "match" substantially alters the sentence's meaning.
\item In the second, the "is dying" implies an ongoing process which is not correct, given the context.
\end{itemize}

\section{Post-editing quality experiment}
\label{sec:post-editing}

To assess the quality of our post-editing process, we ran an experiment with three Slovak native speakers who were not involved in post-editing the automatic translations and are currently either PhD students or already hold a PhD. We sampled 30 samples from test sets that were post-edited and 30 automatic translations (done by DeepL) which were not post-edited. The annotators first annotated the full set of 60 samples (for those that were post-edited we utilized the automatically translated versions) on whether the automatic translation is accurate or not (i.e.\ whether post-editing is necessary). In the second step the annotators labeled each of the post-edited translation based on whether it increased, kept the same, or decreased the translation quality compared to the automatic translation. As we had three annotators, the majority vote has been used to obtain the final label for each labelled sample.

Using this annotated data we can then try to shed some light on some of the posed questions which we explore below.

\textbf{To what extent is post-editing necessary?}

Although our dataset was deliberately created to include 30 samples that were manually chosen by the native speaker to be post-annotated, in this experiment only 15 of them (50\%) were labelled by our annotators as requiring post-editing. The full confusion matrix can be seen below:

\[
\begin{tabular}{|c|c|c|c|}
\hline
 & 0 & 1 & Total\\\hline
\textbf{0} & 29 & 1  & 30\\\hline
\textbf{1} & 16 & 14 & 30\\\hline
\textbf{Total} & 45 & 15 & 60\\\hline
\end{tabular}
\]

As the table shows, in 16 cases the annotators changed the label of a sample (compared to the test set) from ``requiring post-editing'' (1) to ``not requiring post-editing'' category (0) meaning the automatic translation was already accurate, while only in a single example the reverse (automatic translation was not accurate but post-editing was not performed) was observed—the actual inaccuracy in translation in this case was that the translated sentence was not well formatted (it did not start with an uppercase character).

\textbf{What is the impact of post-editing on quality?}

The distribution of the labels as obtained from the majority vote of the annotators can be see in the table below:

\[
\begin{tabular}{|l|c|}
\hline
\textbf{post-edit quality} & \textbf{count}\\\hline
increased & 28\\\hline
neutral   & 1\\\hline
decreased & 1\\\hline
\end{tabular}
\]

The results in the table suggest that post-editing overwhelmingly improves the quality of the translated text, except for two cases: in the case of the neutral label the human post-edit was virtually equivalent to the automatically translated version, while in the case of the decreased quality it appears the native speaker who conducted post-editing made an error and mistankenly repeated the same word twice in the sentence.

\textbf{Summary}

In summary, we can interpret the obtained results as follows:
\begin{itemize}
\item (i) automated translations are largely accurate, as the annotators deemed only 15/60 samples as needing post-editing
\item (ii) when post-edited, quality improved in 28/30 cases, indicating that editing generally renders the translation more accurate
\item (iii) in cases where the original was already acceptable, unnecessary post-editing led to quality degradation in only one instance
\item (iv) among the non-post-edited translations, only one sample was found inaccurate
\end{itemize}

These findings suggest that the automated translation system generally produces high quality translations (also confirmed by the native speakers doing post-editing, as they only did it for less than 5\% of the samples), with selective post-editing providing substantial benefits while rarely causing harm.

\newpage
\section{Guidelines for Assessing Adequacy and Fluency of Translated Text} \label{apx:guidelines}

\section*{Objective}
This section contains the guidelines evaluators should follow to evaluate Adequacy and Fluency in translations. The goal of these guidelines, which contain explanations and examples of each score, is to homogenize the criteria of evaluators to obtain reproducible and reliable results in translation quality evaluation.

\section*{Fluency}
Fluency can be understood as to what extent the translation is \textit{“one that is well-formed grammatically, contains correct spellings, adheres to common use of terms, titles and names, is intuitively acceptable and can be sensibly interpreted by a native speaker”} (Linguistic Data Consortium).

Fluency must be evaluated first, \textbf{before reading the source text}, to evaluate whether the translation reads naturally in the target language. Why? It is worth stressing that a translation may be flawlessly fluent (that is, the translation may be perfectly-formed and follow all the target language rules), but may have adequacy problems (e.g., may contain only partially the meaning of the source sentence). After the evaluation of Fluency, you can then evaluate Adequacy (more details on how to evaluate Adequacy below).

In the table below, Fluency scores are defined and some examples of their evaluation are provided:

\begin{tcolorbox}[colback=gray!5!white, colframe=black, title=SCORE 1 - INCOMPREHENSIBLE, fonttitle=\bfseries, width=\textwidth]
The translation is poorly written and \textbf{nothing can be understood} and/or the grammar and syntactical \textbf{rules of the target language are not respected at all}.

\begin{center}
\begin{tabular}{p{0.45\textwidth} | p{0.45\textwidth}}
\toprule
\textbf{ENGLISH SOURCE TEXT} & \textbf{SLOVAK TRANSLATION} \\
\midrule
Record-high inflation in the Eurozone for October. & Záznam vysoká inflácia Eurozóna za Október. \\
\midrule
Men jailed for life for murder of Sheffield solicitor. & Mužovia uväznení pre životnosť za vraždenie zo Sheffieldskej poradkyňa. \\
\bottomrule
\end{tabular}
\end{center}
\end{tcolorbox}

\begin{tcolorbox}[colback=gray!5!white, colframe=black, title=SCORE 2 - DISFLUENT, fonttitle=\bfseries, width=\textwidth]
The sentence is incorrectly written, difficult to understand, and/or the grammar and syntactical rules of the target language are scarcely respected. \textbf{Less than 50\% of the translation can be understood.}

\begin{center}
\begin{tabular}{p{0.45\textwidth} | p{0.45\textwidth}}
\toprule
\textbf{ENGLISH SOURCE TEXT} & \textbf{SLOVAK TRANSLATION} \\
\midrule
Dishonest solicitor ‘too embarrassed’ to tell bosses of her mistake. & Nečestný advokátka príliš hanba hovoriť šéfov o chyba. \\ 
\midrule
New lord chancellor leads events to mark the opening of the legal year. & Nový lord kancelár vedie udalosti pre značka otváranie rok práva. \\
\bottomrule
\end{tabular}
\end{center}
\end{tcolorbox}

\begin{tcolorbox}[colback=gray!5!white, colframe=black, title=SCORE 3 - GOOD, fonttitle=\bfseries, width=\textwidth]
The translation is partially understood as it is fluent but has minor grammatical or syntactical mistakes. \textbf{More than 50\% of the translation can be understood.}

\begin{center}
\begin{tabular}{p{0.45\textwidth} | p{0.45\textwidth}}
\toprule
\textbf{ENGLISH SOURCE TEXT} & \textbf{SLOVAK TRANSLATION} \\
\midrule
In most cases, the enquiry will not extend beyond an assessment of the parties' needs. & Vo väčšine prípadov sa vyšetrovanie nebude šíriť za hodnotenie potrieb strán. \\ 
\midrule
Drawing a line between marital and non-marital assets is not always easy. & Vytváranie hranice medzi manželskými a nemanželskými majetkami nie je vždy jednoduché. \\
\bottomrule
\end{tabular}
\end{center}
\end{tcolorbox}

\begin{tcolorbox}[colback=gray!5!white, colframe=black, title=SCORE 4 - FLAWLESS, fonttitle=\bfseries, width=\textwidth]
The sentence is \textbf{very fluent and contains no errors}, and the grammar and syntactical \textbf{rules of the target language are completely respected.}

\begin{center}
\begin{tabular}{p{0.45\textwidth} | p{0.45\textwidth}}
\toprule
\textbf{ENGLISH SOURCE TEXT} & \textbf{SLOVAK TRANSLATION} \\
\midrule
Record-high inflation in the Eurozone for October. & Rekordná inflácia v eurozóne v októbri. \\
\midrule
Which nuclear weapons could Putin use against Ukraine? & Ktoré jadrové zbrane by mohol Putin použiť proti Ukrajine? \\
\bottomrule
\end{tabular}
\end{center}
\end{tcolorbox}

\section*{Adequacy}
Adequacy can be understood as \textit{“how much of the meaning expressed in the gold-standard translation or the source is also expressed in the target translation”} (Linguistic Data Consortium).

In the table below, Adequacy scores are defined, with examples provided for each score:

\begin{tcolorbox}[colback=gray!5!white, colframe=black, title=SCORE 1 - NONE, fonttitle=\bfseries, width=\textwidth]
\textbf{None of the meaning} in the source is contained in the translation.

\begin{center}
\begin{tabular}{p{0.45\textwidth} | p{0.45\textwidth}}
\toprule
\textbf{ENGLISH SOURCE TEXT} & \textbf{SLOVAK TRANSLATION} \\
\midrule
Record-high inflation in the Eurozone for October. & EURIBOR dosiahol ročné minimum v poslednom prehľade. \\
\midrule
Ms. Eileen Patterson has been newly appointed as deputy chair of the board. & Pán Eileen Patterson bol práve odvolaný ako pokladník politickej strany. \\
\bottomrule
\end{tabular}
\end{center}
\end{tcolorbox}

\begin{tcolorbox}[colback=gray!5!white, colframe=black, title=SCORE 2 - LITTLE, fonttitle=\bfseries, width=\textwidth]
Small fragments of the meaning in the source are contained in the translation. \textbf{Less than 50\% of the total meaning in the source is contained in the translation.}

\begin{center}
\begin{tabular}{p{0.45\textwidth} | p{0.45\textwidth}}
\toprule
\textbf{ENGLISH SOURCE TEXT} & \textbf{SLOVAK TRANSLATION} \\
\midrule
As a non-executive director, he will contribute to the good governance of the Department of Health. & Ako výkonný riaditeľ prispeje k dobrej správe vlády. \\
\midrule
Our new partners are highly experienced advisers in sectors impacted by the impending departure of the UK from the EU. & Naši noví partneri majú veľa skúseností s problémami nezávislosti medzi Severným Írskom a Írskou republikou. \\
\bottomrule
\end{tabular}
\end{center}
\end{tcolorbox}

\begin{tcolorbox}[colback=gray!5!white, colframe=black, title=SCORE 3 - MOST, fonttitle=\bfseries, width=\textwidth]
Almost all the meaning in the source is contained in the translation. \textbf{More than 50\% of the total meaning of the source is contained in the translation.}

\begin{center}
\begin{tabular}{p{0.45\textwidth} | p{0.45\textwidth}}
\toprule
\textbf{ENGLISH SOURCE TEXT} & \textbf{SLOVAK TRANSLATION} \\
\midrule
Record-high inflation in the Eurozone for October. & Rekordná inflácia v eurozóne. \\
\midrule
Both appointments are for a period of three years, beginning 1 October 2022, with the possibility of an extension to a maximum of six years. & Obe vymenovania platia na obdobie troch rokov, s možnosťou predĺženia na maximálne šesť rokov. \\
\bottomrule
\end{tabular}
\end{center}
\end{tcolorbox}

\begin{tcolorbox}[colback=gray!5!white, colframe=black, title=SCORE 4 - EVERYTHING, fonttitle=\bfseries, width=\textwidth]
All the meaning in the source is contained in the translation, no more, no less.

\begin{center}
\begin{tabular}{p{0.45\textwidth} | p{0.45\textwidth}}
\toprule
\textbf{ENGLISH SOURCE TEXT} & \textbf{SLOVAK TRANSLATION} \\
\midrule
Record-high inflation in the Eurozone for October. & Rekordná inflácia v eurozóne v októbri. \\
\midrule
Which nuclear weapons could Putin use against Ukraine? & Ktoré jadrové zbrane by mohol Putin použiť proti Ukrajine? \\
\bottomrule
\end{tabular}
\end{center}
\end{tcolorbox}

\newpage

\section*{Additional Examples}
This section contains further explanations and examples. These examples have been shown to be problematic after some iterations, so we are adding them here to help you understand how to score each segment if you encounter similar cases.

\begin{tcolorbox}[colback=gray!5!white, colframe=black, title=Fluency is Independent from Adequacy, fonttitle=\bfseries, width=\textwidth]
\textbf{Remember that you should assess Fluency without considering Adequacy.}

\begin{center}
\begin{tabular}{p{0.45\textwidth} | p{0.45\textwidth}}
\toprule
\textbf{ENGLISH SOURCE TEXT} & \textbf{SLOVAK TRANSLATION} \\
\midrule
WHEREAS the \textcolor{red}{SUPPLIER} wishes to engage the \textcolor{red}{AGENT} to sell his products in the territory of Spain; & KEĎŽE si \textcolor{red}{DODÁVATEĽ} želá angažovať \textcolor{red}{DODÁVATEĽA}, aby predával jeho produkty na území Španielska; \\
\bottomrule
\end{tabular}
\end{center}

In the example above, both “SUPPLIER” and “AGENT” are translated as “DODÁVATEĽ,” which is incorrect in terms of \textbf{Adequacy} and should receive a score of \textbf{SCORE 3 - MOST}. \\

Nevertheless, the translation reads naturally in Slovak, so the \textbf{Fluency} score should be \textbf{SCORE 4 - FLAWLESS}.
\end{tcolorbox}

\begin{tcolorbox}[colback=gray!5!white, colframe=black, title=Dealing with Percentages of Appropriateness – Adequacy is Dependent on Fluency, fonttitle=\bfseries, width=\textwidth]
This table explains in detail how to assess the percentage of appropriateness in Fluency and Adequacy based on the +/-50\% threshold. Fluency mistakes are marked in red; Adequacy mistakes are marked in blue. Unlike the previous example, Adequacy here is dependent on Fluency.

\begin{center}
\begin{tabular}{p{0.45\textwidth} | p{0.45\textwidth}}
\toprule
\textbf{ENGLISH SOURCE TEXT} & \textbf{SLOVAK TRANSLATION} \\
\midrule
\textcolor{red}{The AGENT shall in and about the execution of his activity make} every effort to safeguard \textcolor{red}{the interests} of the SUPPLIER, \textcolor{red}{in conformity with} the best business practice. & \textcolor{red}{Agent by mal vo vykonávanie jeho} činností robiť všetko možné pre ochranu \textcolor{red}{záujmy dodávateľ}, \textcolor{red}{v zhoda} s najlepšou obchodnou praxou. \\

The AGENT shall in and about the execution of his activity make \textcolor{blue}{every effort to safeguard the interests of the SUPPLIER}, \textcolor{blue}{in conformity} with the best business practice. & Agent by mal v rámci svojej činnosti urobiť \textcolor{blue}{všetko pre obranu distribútora}, \textcolor{blue}{napriek nezhode} s najlepšou obchodnou praxou. \\

\bottomrule
\end{tabular}
\end{center}

In terms of \textcolor{red}{Fluency (red)}, more than 50\% of the translation has fluency problems. Therefore, the appropriate Fluency score is \textbf{SCORE 2 - DISFLUENT} because the translation does not flow naturally in Slovak.

In terms of \textcolor{blue}{Adequacy (blue)}, most of the meaning is present but lacks correct wording, impacting the Adequacy score. As a result, the appropriate Adequacy score would be \textbf{SCORE 2 - LITTLE}.
\end{tcolorbox}

\begin{tcolorbox}[colback=yellow!5!white, colframe=black, title=Important Note, fonttitle=\bfseries, width=\textwidth]
Though there may be translations that are fluent but not adequate, for a translation to be adequate, it must also be fluent. Thus, an adequate translation should also be fluent.
\end{tcolorbox}

\end{document}